\documentclass{article} %
\usepackage[final]{colm2026_conference}

\usepackage[table]{xcolor}
\usepackage[T1]{fontenc}
\usepackage{microtype}
\usepackage{hyperref}
\usepackage{xurl}
\usepackage{booktabs}

\usepackage{graphicx}
\usepackage{multirow}
\usepackage{amssymb}
\usepackage{pifont}
\newcommand{\xmark}{\ding{55}}

\usepackage{nicematrix}
\usepackage{amsmath}
\usepackage{enumitem}
\usepackage{bbm}
\usepackage{caption}
\usepackage{subcaption}
\usepackage{wrapfig}
\usepackage{tcolorbox}
\usepackage{multicol}
\usepackage{framed, algorithmic}
\tcbuselibrary{breakable}
\usepackage{nicefrac}
\usepackage[normalem]{ulem}
\usepackage{fvextra}
\usepackage{tikz}
\usepackage{float}
\usetikzlibrary{arrows.meta,calc,positioning}

\usepackage{xspace}
\newcommand{\ours}{AggAgent\xspace}
\newcommand{\MV}{MV\xspace}
\newcommand{\BoN}{BoN\xspace}
\newcommand{\WMV}{WMV\xspace}
\newcommand{\FewTool}{FewTool\xspace}
\newcommand{\SolAgg}{SolAgg\xspace}
\newcommand{\SummAgg}{SummAgg\xspace}

\definecolor{LightOrange}{RGB}{255,245,235}
\tcbuselibrary{breakable,skins,listings}

\lstdefinelanguage{json}{
  basicstyle=\ttfamily\footnotesize,
  breaklines=true,
  showstringspaces=false,
  morestring=[b]",
}

\usepackage{lineno}

\definecolor{darkblue}{rgb}{0, 0, 0.5}
\definecolor{customgreen}{HTML}{E6F1E2}
\definecolor{customblue}{HTML}{E2ECFF}
\hypersetup{colorlinks=true, citecolor=darkblue, linkcolor=darkblue, urlcolor=darkblue}

\title{Agentic Aggregation for Parallel Scaling of Long-Horizon Agentic Tasks}

\author{Yoonsang Lee $\;$ Howard Yen $\;$ Xi Ye $\;$ Danqi Chen \\ 
Princeton Language and Intelligence, Princeton University \\
\texttt{\{yoonsang, hyen, danqic\}@cs.princeton.edu} $\;$ xi.ye@princeton.edu  }

\begin{document}

\ifcolmsubmission
\linenumbers
\fi

\maketitle

\begin{abstract}
We study parallel test-time scaling for long-horizon agentic tasks such as agentic search and deep research, where multiple rollouts are generated in parallel and aggregated into a final response. While such scaling has proven effective for chain-of-thought reasoning, agentic tasks pose unique challenges: trajectories are long, multi-turn, and tool-augmented, and outputs are often open-ended. Aggregating only final answers discards rich information from trajectories, while concatenating all trajectories exceeds the model's context window. To address this, we propose \textbf{\ours}, an \emph{aggregation agent that treats parallel trajectories as an environment}. We equip it with lightweight tools to inspect candidate solutions and search across trajectories, enabling it to navigate and synthesize information on demand. Across six benchmarks and three model families (GLM-4.7, Qwen3.5, MiniMax-M2.5), \ours outperforms all existing aggregation methods---by up to 5.3\% absolute on average and 10.3\% on two deep research tasks---while adding minimal overhead, as the aggregation cost remains bounded by a single agentic rollout. Our findings establish agentic aggregation as an effective and cost-efficient approach to parallel test-time scaling.\footnote{Our code is available at \url{https://github.com/princeton-pli/AggAgent}.}

\end{abstract}

\section{Introduction}

Scaling test-time compute has emerged as a promising avenue for enhancing the performance of large language models (LLMs)~\citep{wei2022chain, wang2022self, brown2024large, welleck2024from,snell2025scaling,muennighoff2025s1}. This success has largely been demonstrated in standard chain-of-thought (CoT) tasks such as mathematical reasoning and coding~\citep{qi2025learning, zhao2025majority}. However, \textit{long-horizon agentic tasks} 

\begin{figure}[H]
    \centering
    \vspace{-10pt}
    \includegraphics[width=\linewidth]{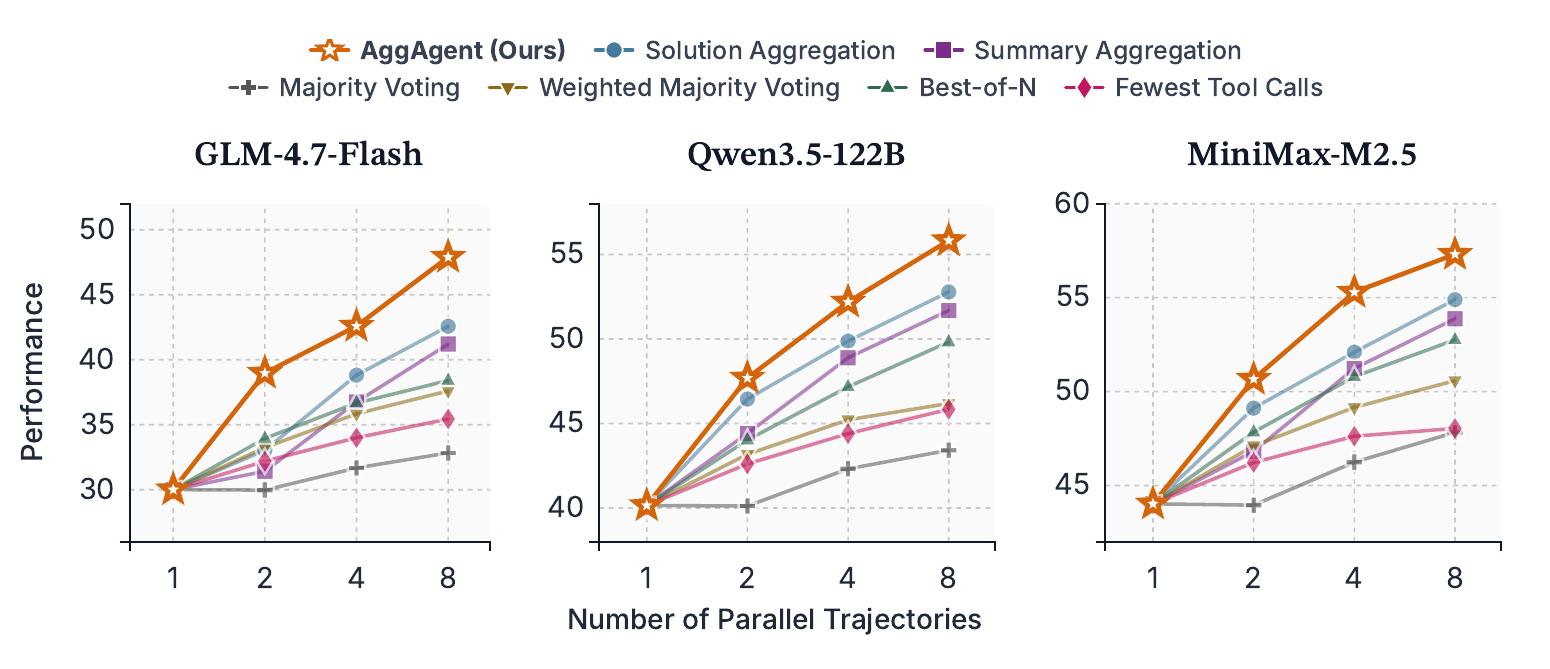}
    \vspace{-20pt}
    \caption{\textbf{\ours consistently outperforms existing aggregation methods.} We measure the average performance across six long-horizon agentic benchmarks (Section~\ref{sec:tasks}) against the number of parallel trajectories. The same model as the rollout agent serves as the aggregator.
    }
    \label{fig:average_k}
\end{figure}

(e.g., deep research, software engineering, web navigation) present a fundamentally different challenge: trajectories are multi-turn, spanning hundreds of steps with interleaved tool calls and observations~\citep{yao2022react}. 

In this work, we study \textit{parallel scaling} for long-horizon agentic tasks, an approach proven effective for CoT reasoning tasks yet largely unexplored for agentic search and deep research. Parallel scaling enables simultaneous generation of multiple independent trajectories, offering a natural computational advantage~\citep{zhao2025majority}. Furthermore, parallel scaling offers substantial improvements over single-agent performance~\citep{wei2025browsecomp, li2025openresearcher}: for instance, GLM-4.7-Flash~\citep{zeng2025glm} improves from 27\% to 59\% on BrowseComp~\citep{wei2025browsecomp} and 25\% to 51\% on HLE~\citep{phan2025humanity} when scaling from Pass@1 to Pass@8, confirming that correct solutions frequently exist within the parallel rollouts. 

The central question then becomes how to effectively \textit{aggregate} these trajectories. While voting and solution aggregation methods have proven effective for mathematical reasoning and coding~\citep{wang2022self, fu2026deep,zhao2025majority}, long-horizon agentic tasks pose distinct challenges: evidence is sparse and distributed across multi-turn trajectories, demanding reasoning far beyond shallow heuristics and final solutions~\citep{li2025parallelmuse}. Furthermore, individual trajectories may each capture only partial progress toward a complex task, requiring cross-trajectory synthesis to assemble a complete solution~\citep{chang2026karl}. However, existing approaches fall short (Table~\ref{tab:baselines}): aggregating only final solutions discards the rich information in trajectories, summarizing trajectories is expensive and lossy, and concatenating all trajectories into a single context is infeasible, as each trajectory spans hundreds of thousands of tokens.

\begin{figure}[tbp]
    \centering
    \includegraphics[width=\linewidth, height=0.38\textheight, keepaspectratio]{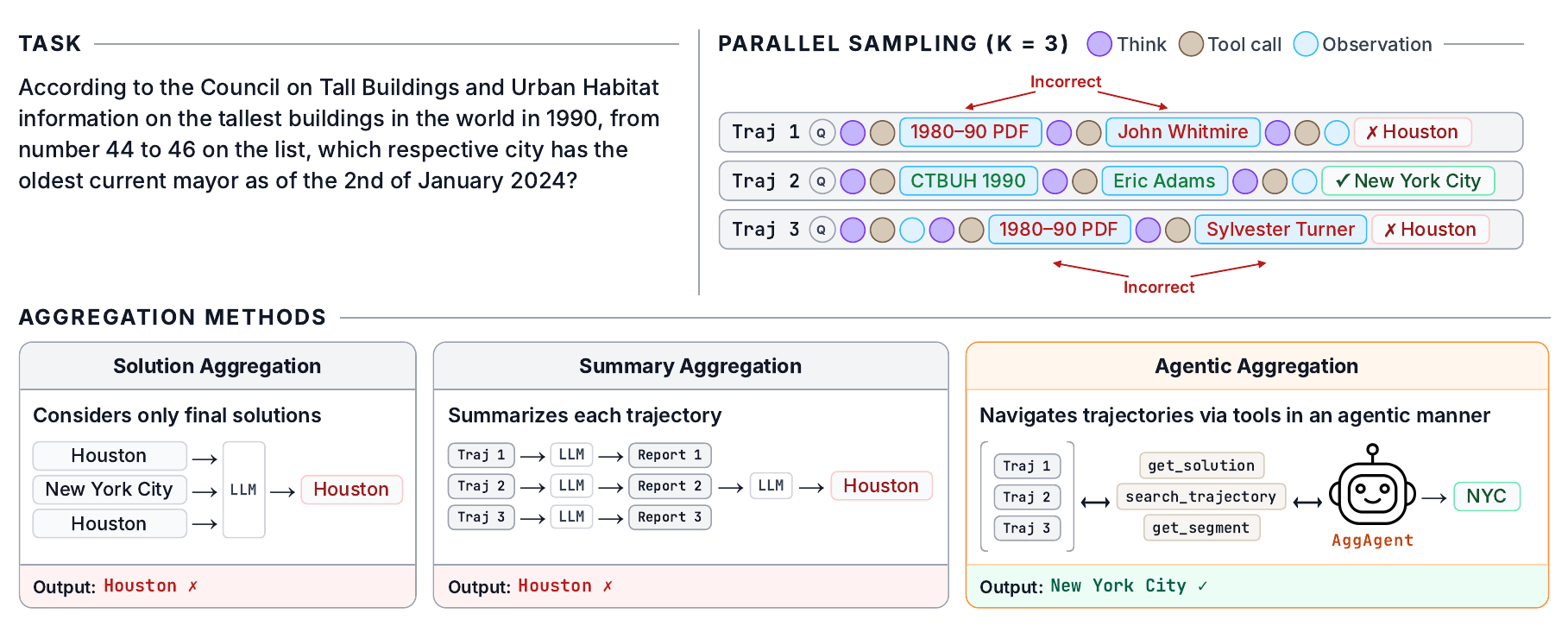}
    \caption{\textbf{Overview of aggregation methods for parallel scaling.} \textbf{(Top)} An agent produces $K=3$ independent rollouts on a long-horizon task. \textbf{(Bottom)} \emph{Solution Aggregation} feeds only final solutions to an LLM, discarding intermediate reasoning. \emph{Summary Aggregation} compresses each trajectory into a lossy summary. \textbf{\ours} (ours) navigates trajectories via tools in an agentic manner, enabling full-fidelity cross-trajectory reasoning at low cost.
    }
    \label{fig:overview}
\end{figure}

To address these challenges, we propose \textbf{\ours}, which frames aggregation itself as an agentic task, treating the set of trajectories as an environment to interact with (Figure~\ref{fig:overview}). \ours interacts with trajectories on demand through three lightweight tools: \texttt{get\_solution} (retrieve final solutions of one or all trajectories), \texttt{search\_trajectory} (keyword search within a trajectory), and \texttt{get\_segment} (read a specific range of steps). Since these tools operate entirely over an array of completed trajectories---rather than invoking external services like web search or code execution---they incur none of the API cost or latency of such external tools. This agentic aggregation allows for cross-trajectory reasoning and synthesis without the cost of loading all trajectories into context, preserving full fidelity while keeping the aggregation cost bounded by a single agentic rollout.

We evaluate \ours on six agentic search and deep research benchmarks across three model families (GLM-4.7, Qwen3.5, MiniMax-M2.5). \ours consistently outperforms all baselines (Figure~\ref{fig:average_k}), improving performance by up to 5.3 points on average and 10.3 points on deep research tasks. Furthermore, \ours is Pareto-optimal in cost and performance (Figure~\ref{fig:eff_mini}), adding only a 5.7\% overhead over the cost of running 8 parallel agents---compared to 41\% for Summary Aggregation---by selectively reading portions of trajectories rather than loading them entirely into context. Notably, \ours even surpasses Pass@8 (Section~\ref{sec:stronger}), demonstrating that effective aggregation can synthesize correct solutions beyond what any single rollout achieves. To summarize, our results show that agentic aggregation is a scalable and effective paradigm for test-time scaling in long-horizon tasks. We demonstrate that \ours is effective with off-the-shelf LLMs, and our framework further opens a promising direction for training aggregator agents.

\section{Problem Formulation}

\begin{table}[tbp]
\centering
\small
\setlength{\tabcolsep}{5pt}
\begin{tabular}{l|cccccc|c}
\toprule
\textbf{Criterion} & \textbf{\MV} & \textbf{\WMV} & \textbf{\BoN} & \textbf{\FewTool} & \textbf{\SolAgg} & \textbf{\SummAgg} & \textbf{\ours (Ours)} \\
\midrule
\textbf{Task-agnostic}    & {\color{red}\xmark} & {\color{red}\xmark} & {\color{green!60!black}\ding{51}} & {\color{green!60!black}\ding{51}} & {\color{green!60!black}\ding{51}} & {\color{green!60!black}\ding{51}} & {\color{green!60!black}\ding{51}} \\
\textbf{Non-heuristic}    & {\color{red}\xmark} & {\color{red}\xmark} & {\color{red}\xmark} & {\color{red}\xmark} & {\color{green!60!black}\ding{51}} & {\color{green!60!black}\ding{51}} & {\color{green!60!black}\ding{51}} \\
\textbf{Trajectory Info.} & {\color{red}\xmark} & {\color{red}\xmark} & {\color{red}\xmark} & {\color{red}\xmark} & {\color{red}\xmark}     & {\color{green!60!black}\ding{51}} & {\color{green!60!black}\ding{51}} \\
\textbf{Full Fidelity}    & {\color{red}\xmark} & {\color{red}\xmark} & {\color{red}\xmark} & {\color{red}\xmark} & {\color{red}\xmark}     & {\color{red}\xmark}     & {\color{green!60!black}\ding{51}} \\
\textbf{Aggregation Cost} & Zero   & Zero   & Zero   & Zero   & Low        & High       & Low \\
\bottomrule
\end{tabular}
\caption{\textbf{Comparison of aggregation methods}. \textbf{Task-agnostic}: not restricted to single short answers, applicable to multi-answer and long-form tasks. \textbf{Non-heuristic}: does not rely on shallow signals such as frequency or confidence. \textbf{Trajectory Info.}: reasons over full trajectories beyond final solutions. \textbf{Full Fidelity}: accesses trajectories without compression or information loss. \textbf{Aggregation Cost}: additional inference cost beyond the rollouts.}
\vspace{-7pt}
\label{tab:baselines}
\end{table}

We study long-horizon tasks such as agentic search and deep research. Given a problem $q$, the goal is to produce an output $y$, which may be either a short answer, a set of answers, or a long-form report. An agent $A$ interacts with an external environment to solve $q$ by generating a trajectory $T = (q, r_1, a_1, o_1, \dots, r_m, a_m, o_m, y)$, where each step consists of internal thinking $r_j$, tool call $a_j$, and resulting observation $o_j$. In parallel scaling, we run the same agent independently for $K$ times, yielding $\mathcal{T} = \{T_1, T_2, \dots, T_K\}$. This setting is natural in production deployments of frontier systems~\citep{AnthropicDeepResearch2025,Cursor2025,team2026kimi}, as the latency is determined by the slowest of the $K$ rollouts.

The core problem centers around how to aggregate these $K$ trajectories and synthesize a better solution $y$. We formalize aggregation as a function $f: (q, \mathcal{T}) \mapsto \hat{y}$, where $\hat{y}$ is the aggregated final solution. For agentic search, $\hat{y}$ is evaluated against the ground-truth answer $y^*$, while for deep research, it is evaluated against problem-specific rubrics.

\paragraph{Existing methods}
We adopt the following aggregation methods from prior works:
\begin{itemize} 
[leftmargin=*,noitemsep,nolistsep,itemsep=3pt]
    \item \textbf{Majority Voting} (\MV)~\citep{wang2022self}: selects the most frequent solution. However, voting methods do not work for multi-answer or long-form generation tasks.
    \item \textbf{Best-of-N} (\BoN)~\citep{cobbe2021training,uesato2022solving}: selects the solution with the highest self-reported confidence. While reward model scores are commonly used as the weight for CoT reasoning tasks, recent works on agentic tasks have adopted the model's self-reported confidence as the weight~\citep{wei2025browsecomp, zhu2026re}.
    \item \textbf{Weighted Majority Voting} (\WMV)~\citep{li-etal-2023-making,wu2024scaling}: selects the solution with the highest total weight, where each solution is weighted by self-reported confidence.
    \item \textbf{Fewest Tool Calls} (\FewTool)~\citep{liu2025deepseek, lu2025deepdive}: selects the solution from the trajectory that required the fewest tool calls.
    \item \textbf{Solution Aggregation} (\SolAgg)~\citep{qi2025learning,zhao2025majority,qiao2025webresearcher}: concatenates all $K$ solutions and prompts an LLM to synthesize a final solution.
    \item \textbf{Summary Aggregation} (\SummAgg)~\citep{li2025parallelmuse, hu2026pacore}: compresses each trajectory into a summary report, concatenates all reports, and prompts an LLM to generate a final solution. 
\end{itemize}

We classify these baselines into two groups. \textit{Heuristic methods} (\MV, \WMV, \BoN, \FewTool) rely on shallow signals such as answer frequency, confidence scores, or trajectory length, without inspecting trajectory content. \textit{LLM-based methods} (\SolAgg, \SummAgg) leverage LLM reasoning for solution synthesis but remain limited: Solution Aggregation aggregates only final solutions, discarding all intermediate evidence, while Summary Aggregation compresses each trajectory into a summary report before aggregation, incurring irreversible information loss and requiring $K$ additional LLM calls. Table~\ref{tab:baselines} summarizes these limitations.

\section{Our Approach: {\ours}}
To address the aforementioned limitations, we propose \ours, where the aggregator interacts with $\mathcal{T}$ directly in an agentic manner. The initial user message provides the problem $q$ and trajectory metadata, which consists of the number of steps, total token count, and tool usage statistics for each trajectory $T_i$. The trajectories themselves are not pre-loaded into the context but reside in the environment and are retrieved on-demand, keeping the aggregation cost bounded by a single context window independent of $K$. This design enables cross-trajectory reasoning at full fidelity and low cost (Table~\ref{tab:baselines}).

\paragraph{Tool design} We provide \ours with four tools as follows\footnote{We provide the full tool descriptions in Figure~\ref{fig:aggagent_tools} in the appendix.}:
\begin{itemize}
[leftmargin=*,noitemsep,nolistsep,itemsep=3pt]
    \item \textbf{\texttt{get\_solution(traj\_id)}}: Retrieves the final solution from each trajectory's last step. Unless {traj\_id} is specified, returns all $K$ solutions by default.
    \item \textbf{\texttt{search\_trajectory(traj\_id, query, role, k)}}: Searches for keywords within a single trajectory. Returns the top matching $k$ steps ranked by ROUGE-L score~\citep{lin-2004-rouge}.
    \item \textbf{\texttt{get\_segment(traj\_id, start\_step, end\_step)}}: Reads the full content of a contiguous range of steps from a single trajectory. Returns raw thinking and tool observations for the specified window.
    \item \textbf{\texttt{finish()}}: Submits the final {solution} along with a {reason} for the aggregation.
\end{itemize}

\paragraph{Workflow} The tool design reflects a natural coarse-to-fine investigative workflow. Rather than reading every trajectory in full, the agent first surveys the metadata and solutions loaded by \texttt{get\_solution} to identify consensus and disagreements across the rollouts, and to pinpoint trajectories worth closer inspection. It then selectively dives into individual trajectories via \texttt{search\_trajectory} and \texttt{get\_segment} to verify key claims against thinking blocks and tool observations. After sufficient cross-trajectory validation, it calls \texttt{finish} to submit the final synthesized solution. We provide the full prompts in Appendix~\ref{app:prompts_aggagent}. 

\paragraph{Cost analysis}
Aggregation cost is measured in additional LLM calls beyond the $K$ rollouts. Heuristic methods require zero additional calls, while Solution Aggregation requires a single LLM call to synthesize the $K$ solution candidates. Summary Aggregation is the most expensive, as it requires $K$ separate LLM calls before aggregation, each of which can be as long as the maximum context window. In contrast, \ours's total context is bounded by a single context window of the model, keeping aggregation cost independent of $K$ and comparable to Solution Aggregation. Moreover, as \ours's tools operate entirely over the in-memory trajectory array, they incur none of the latency or cost of external tools like web search and code execution.

\section{Experimental Setup}

\paragraph{Tasks}
\label{sec:tasks}
We evaluate on six long-horizon agentic tasks broadly in two categories: 1) \textbf{Agentic search}: \textbf{BrowseComp}~\citep{wei2025browsecomp} consists of challenging factual questions demanding exhaustive multi-step web browsing; \textbf{BrowseComp-Plus}~\citep{chen2025browsecomp} enables controlled evaluation on BrowseComp questions by replacing web search with a local knowledge base; \textbf{HLE}~\citep{phan2025humanity} covers expert-level questions across diverse academic disciplines, emphasizing rigorous reasoning; \textbf{DeepSearchQA}~\citep{gupta2026deepsearchqa} targets multi-answer queries where completeness across all valid answers is required. 2) \textbf{Deep research}: \textbf{Healthbench-Hard}~\citep{arora2025healthbench} requires generating comprehensive long-form responses to challenging medical queries; \textbf{ResearchRubrics}~\citep{sharma2025researchrubrics} poses open-ended research tasks evaluated against detailed, multi-criterion rubrics.

Due to high costs of long-horizon rollouts, we randomly sample subsets for evaluation\footnote{We report standard deviation in Appendix~\ref{app:main_results}.}. For HLE, we use 155 search-focused questions~\citep{li2025parallelmuse}. For ResearchRubrics, we use the full 101 problems. For Healthbench-Hard, we randomly sample 100 instances. For all other tasks, we randomly sample 150 questions following \citet{sun2025scaling, zhang2025recursive}.

\paragraph{Models}
We employ three model families of varying sizes for base trajectory rollouts: GLM-4.7-Flash (30B)~\citep{zeng2025glm}, Qwen3.5-122B-A10B (122B)~\citep{qwen3.5}, and MiniMax-M2.5 (229B)~\citep{MiniMax2026}. We adopt Tongyi DeepResearch~\citep{team2025tongyi} as the agent scaffold, use native function calling, allow at most 128K context length with 100 tool calls, and sample 8 independent trajectories. For LLM-based aggregation, we use the same model as the rollout agent.

\paragraph{Rollout agent tools} For BrowseComp-Plus, we follow the official implementation and provide \texttt{search} and \texttt{get\_document} tools using Qwen3-Embedding-8B~\citep{zhang2025qwen3} as the retriever. For all other tasks, we provide \texttt{search} and \texttt{visit} tools following prior work~\citep{yen2025lost, li2025websailor, wu2025resum, gao2025beyond}, where \texttt{search} uses Serper API\footnote{\url{https://serper.dev}} for Google search and \texttt{visit} uses crawl4ai\footnote{\url{https://github.com/unclecode/crawl4ai}} to scrape a specific web page. 

\paragraph{Cost and latency}
We report total cost and latency per query. Cost includes rollout, tool calls, and aggregation, while latency includes rollout and aggregation.

\paragraph{Evaluation}
We employ LLM-as-a-judge, following the official evaluation setting. Specifically, we use Qwen3.5-397B-A17B for ResearchRubrics and GPT-4.1~\citep{achiam2023gpt} for all other datasets. We perform bootstrapped sampling for calculating Metric@$K\in\{1,2,4,8\}$. For BrowseComp, BrowseComp-Plus, and HLE, the predicted answer $\hat{y}$ is evaluated against the gold answer $y^*$. For DeepSearchQA, the predicted answer set is considered correct if it exactly matches the gold answer set. For Healthbench-Hard and ResearchRubrics, the long-form response is evaluated across multiple problem-specific rubrics, and the scores are averaged. The rubrics contain negative scores, hence the total score could also be negative.

We provide further experimental details in Appendix~\ref{app:exp}.

\section{Results}

\begin{table}[tbp]
\centering
\resizebox{\linewidth}{!}{%
\begin{NiceTabular}{cl|*{8}{w{c}{1.5cm}}}
\toprule
& \Block{2-1}{\textbf{Task}} & \Block{2-1}{\textbf{Pass@1}} & \multicolumn{4}{c}{\textbf{Heuristic Aggregation@8}} & \multicolumn{3}{c}{\textbf{LLM-based Aggregation@8}} \\
\cmidrule(lr){4-7}\cmidrule(lr){8-10}
& & & \textbf{\MV} & \textbf{\WMV} & \textbf{\BoN} & \textbf{\FewTool} & \textbf{\SolAgg} & \textbf{\SummAgg} & \textbf{\ours} \\
\midrule\midrule

\Block[fill=red!8]{7-1}{\rotatebox{90}{\textbf{GLM-4.7-Flash}}}
& BrowseComp       & 27.42 & 32.67 & 50.67 & 51.33 & 41.33 & 53.33 & 55.33 & \cellcolor{orange!8}\textbf{56.00} \\
& BrowseComp-Plus  & 49.08 & 59.33 & 68.67 & 68.67 & 62.00 & 70.67 & 70.67 & \cellcolor{orange!8}\textbf{71.33} \\
& HLE              & 25.00 & 26.45 & 27.74 & 30.97 & 27.10 & 32.90 & 34.84 & \cellcolor{orange!8}\textbf{37.42} \\
& DeepSearchQA     & 32.42 & -- & -- & 35.33 & 33.33 & 46.00 & 47.33 & \cellcolor{orange!8}\textbf{49.33} \\
& Healthbench-Hard & 8.67  & -- & --  & 9.91  & 8.90  & 15.72 & 7.35  & \cellcolor{orange!8}\textbf{27.99} \\
& ResearchRubrics  & 37.47 & -- & -- & 37.70 & 35.21 & 36.84 & 31.72 & \cellcolor{orange!8}\textbf{45.31} \\
\cmidrule(l){2-10}
& \cellcolor{gray!12}\textbf{Average} & \cellcolor{gray!12}30.01 & \cellcolor{gray!12}32.84 & \cellcolor{gray!12}37.61 & \cellcolor{gray!12}38.99 & \cellcolor{gray!12}34.65 & \cellcolor{gray!12}42.58 & \cellcolor{gray!12}41.21 & \cellcolor{gray!12}\textbf{47.90} \\

\midrule

\Block[fill=customgreen]{7-1}{\rotatebox{90}{\textbf{Qwen3.5-122B}}}
& BrowseComp       & 40.25 & 47.33 & 60.00 & 58.67 & 56.00 & 61.33 & 62.00 & \cellcolor{orange!8}\textbf{62.67} \\
& BrowseComp-Plus  & 58.67 & 66.67 & 69.33 & 74.00 & 70.67 & 73.33 & 73.33 & \cellcolor{orange!8}\textbf{74.67} \\
& HLE              & 39.35 & 43.87 & 45.16 & 53.55 & 41.29 & 50.97 & 50.32 & \cellcolor{orange!8}\textbf{54.19} \\
& DeepSearchQA     & 49.25 & -- & -- & 57.33 & 54.67 & 62.67 & 64.00 &\cellcolor{orange!8}\textbf{66.00} \\
& Healthbench-Hard & 12.87 & -- & -- & 13.01 & 12.83 & 26.30 & 23.00 & \cellcolor{orange!8}\textbf{28.06} \\
& ResearchRubrics  & 40.50 & -- & -- & 42.37 & 39.58 & 42.10 & 37.47 & \cellcolor{orange!8}\textbf{49.36} \\
\cmidrule(l){2-10}
& \cellcolor{gray!12}\textbf{Average} & \cellcolor{gray!12}40.15 & \cellcolor{gray!12}43.42 & \cellcolor{gray!12}46.19 & \cellcolor{gray!12}49.82 & \cellcolor{gray!12}45.84 & \cellcolor{gray!12}52.78 & \cellcolor{gray!12}51.69 & \cellcolor{gray!12}\textbf{55.83} \\

\midrule

\Block[fill=customblue]{7-1}{\rotatebox{90}{\textbf{MiniMax-M2.5}}}
& BrowseComp       & 50.17 & 60.67 & 66.67 & 70.67 & 67.33 & 70.67 & 70.00 & \cellcolor{orange!8}\textbf{71.33} \\
& BrowseComp-Plus  & 64.17 & 68.67 & 75.33 & 76.00 & 74.67 & 76.67 & 76.67 & \cellcolor{orange!8}\textbf{77.33} \\
& HLE              & 45.73 & 53.55 & 57.42 & 58.06 & 46.45 & 54.19 & 58.06 & \cellcolor{orange!8}\textbf{60.00} \\
& DeepSearchQA     & 54.42 & -- & -- & 64.00 & 56.00 & 62.00 & 61.33 & \cellcolor{orange!8}\textbf{65.33} \\
& Healthbench-Hard & 9.67  & --  & --  & 8.79  & 5.34  & 21.84 & 16.92 & \cellcolor{orange!8}\textbf{24.46} \\
& ResearchRubrics  & 39.97 & -- & -- & 39.00 & 38.44 & 44.00 & 40.29 & \cellcolor{orange!8}\textbf{45.42} \\
\cmidrule(l){2-10}
& \cellcolor{gray!12}\textbf{Average} & \cellcolor{gray!12}44.02 & \cellcolor{gray!12}47.83 & \cellcolor{gray!12}50.58 & \cellcolor{gray!12}52.75 & \cellcolor{gray!12}48.04 & \cellcolor{gray!12}54.90 & \cellcolor{gray!12}53.88 & \cellcolor{gray!12}\textbf{57.31} \\

\bottomrule
\end{NiceTabular}%
}
\caption{\textbf{\ours performs the best across all settings.} We evaluate on six long-horizon tasks using three model families, each generating $K{=}8$ independent rollouts. For LLM-based aggregation methods, the same model is used for both rollout and aggregation. Best metric per row is bolded. -- denotes that voting-based methods are not applicable, and we use Pass@1 instead to calculate the average score.}
\label{tab:main_mini}
\end{table}

\begin{figure}[h]
    \centering
    \includegraphics[width=\linewidth]{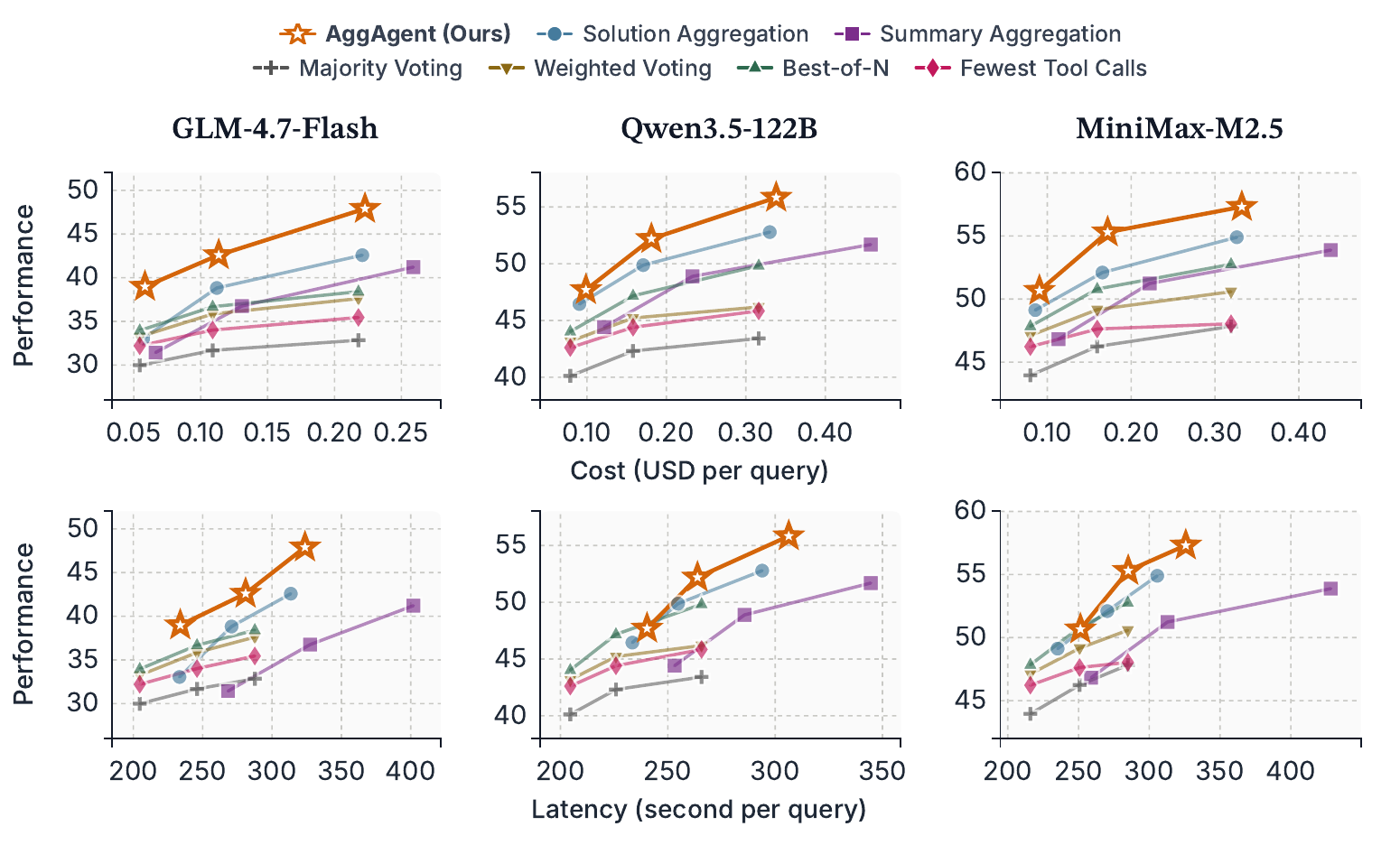}
    \caption{\textbf{\ours achieves a Pareto-optimal performance--efficiency tradeoff}. We compare aggregation methods at $K\in\{2,4,8\}$ parallel samples, averaging performance across six benchmarks, with the same model serving as both rollout agent and aggregator. For each model, the top chart plots average cost (USD per query) vs.\ performance and the bottom chart plots average latency (seconds per query) vs.\ performance.}
    \label{fig:eff_mini}
\end{figure}

\subsection{Main Results}
\label{sec:main_results}
We present the main results on $K=8$ in Table~\ref{tab:main_mini}, and present the full results ($K\in\{1,2,4,8\}$) in Appendix~\ref{app:main_results}. \ours consistently outperforms all baselines across all three models, improving over Pass@1 by 13.3--17.9 points and over the strongest baseline (\SolAgg) by 2.4--5.3 points on average. 

Among the heuristic baselines, Majority Voting performs the worst, while confidence-based methods (\WMV, \BoN) yield substantial improvements, consistent with prior observations~\citep{wei2025browsecomp}. However, the gains are less pronounced on DeepSearchQA, Healthbench-Hard, and ResearchRubrics, where the model is poorly calibrated (see Appendix~\ref{sec:calib} for details). Notably, voting methods are not applicable to multi-answer or long-form tasks, so their performance gains are limited to single-answer tasks only.\footnote{We use Pass@1 instead to calculate the average score.} Fewest Tool Calls has been popular on BrowseComp due to its simplicity~\citep{liu2025deepseek,lu2025deepdive}, but our results show that it also does not generalize well beyond such tasks.

Meanwhile, LLM-based methods consistently outperform heuristic baselines. Summary Aggregation outperforms all other baselines on agentic search tasks, as access to full trajectory evidence aids answer verification. However, it falls significantly behind Solution Aggregation on deep research tasks, as trajectory compression harms the detail and coherence of long-form responses. While no single baseline is universally optimal, \ours achieves strong performance by enabling fine-grained reasoning over trajectories without sacrificing detail or coherence.

\subsection{Cost and Latency vs. Performance}
\label{sec:cost}
We compare cost and latency versus performance for different numbers of samples $K \in \{2, 4, 8\}$ in Figure~\ref{fig:eff_mini} (full breakdown in Appendix~\ref{app:cost_results}). As $K$ increases, all methods become more expensive and slower due to additional rollouts. \ours and Solution Aggregation introduce only minimal overhead beyond heuristic baselines---at $K{=}8$, their overhead over the rollout cost is 5.7\% and 3.7\%, respectively---yet substantially outperform them. Interestingly, their overhead does not increase proportionally to $K$, suggesting that having more samples can sometimes make aggregation easier by providing stronger supporting evidence, partially offsetting the extra computation spent on harder instances. In contrast, Summary Aggregation incurs 41\% overhead at $K{=}8$ because it must compress each trajectory before aggregation. Overall, \ours achieves Pareto-optimal performance and efficiency across benchmarks, showing that agentic aggregation scales efficiently with test-time compute without incurring substantial additional cost.

\section{Analysis and Discussion}

\begin{figure}[b!]
    \centering
    \includegraphics[width=\linewidth]{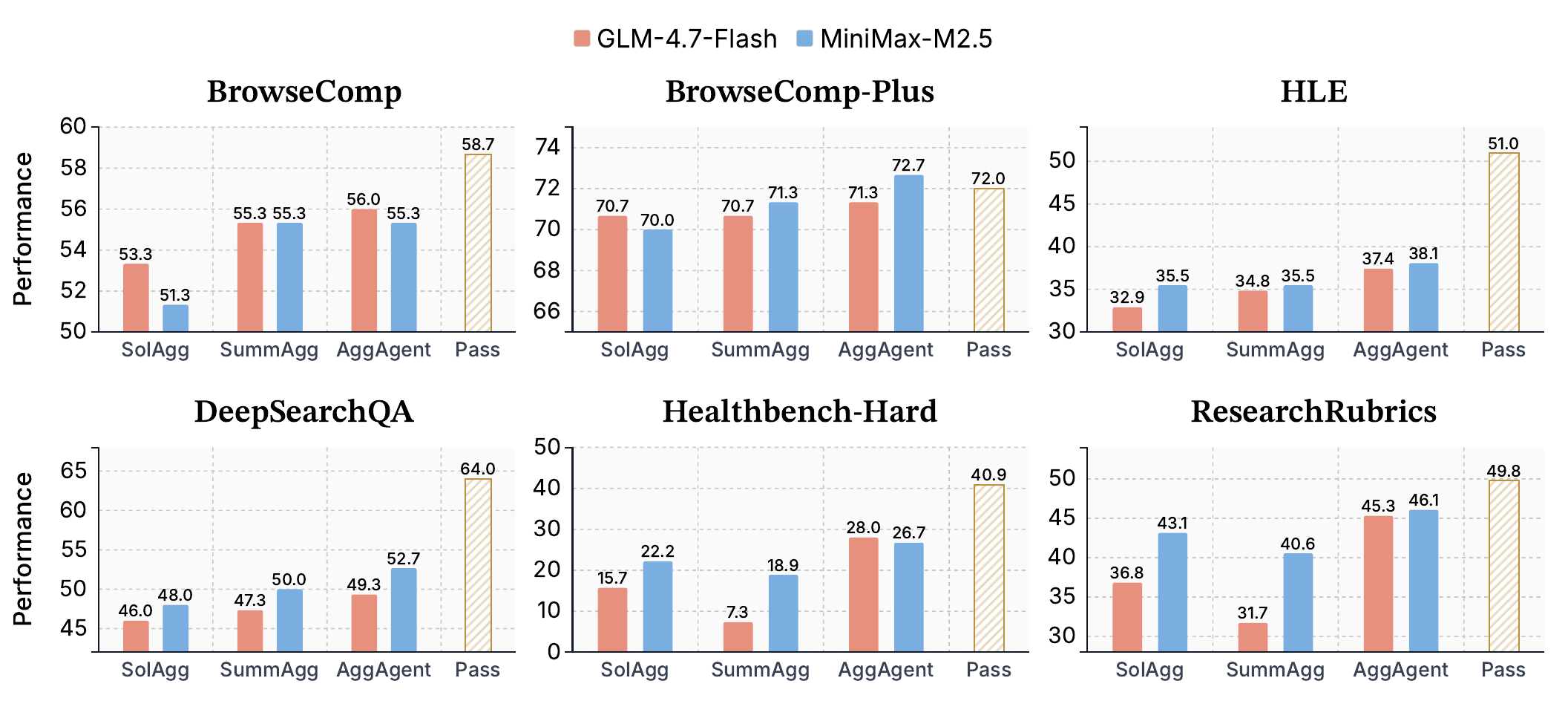}
    \caption{\textbf{Employing a stronger aggregator improves performance on LLM-based aggregation methods}. In all cases, GLM-4.7-Flash serves as the rollout agent; blue bars replace the aggregator with the stronger MiniMax-M2.5, while red bars use GLM-4.7-Flash for both roles. Yellow hatched bars denote Pass@$K$. All methods are evaluated at $K{=}8$. 
    }
    \label{fig:minimax_mini}
\end{figure}

\subsection{Stronger Models for Aggregation}
\label{sec:stronger}

Could aggregation benefit from employing a stronger model? We investigate this by replacing the aggregator with MiniMax-M2.5 while keeping GLM-4.7-Flash as the base rollout agent. As shown in Figure~\ref{fig:minimax_mini} (full results in Table~\ref{tab:minimax} in the appendix), using a stronger aggregator generally improves performance, with MiniMax-based \ours achieving the highest average score and even surpassing Pass@8 on BrowseComp-Plus. This points to \textit{asymmetric model allocation}---a stronger model for aggregation, weaker models for parallel rollouts---as a practical strategy for designing multi-agent systems, consistent with prior work employing stronger models as orchestrators over multiple weaker subagents~\citep{qiao2025webresearcher, zhang2025recursive, akay2026spd}. In our setting, the aggregator plays the orchestrating role, reasoning across parallel trajectories to produce a final solution.

\begin{figure}[t]
    \centering
    \includegraphics[width=\linewidth]{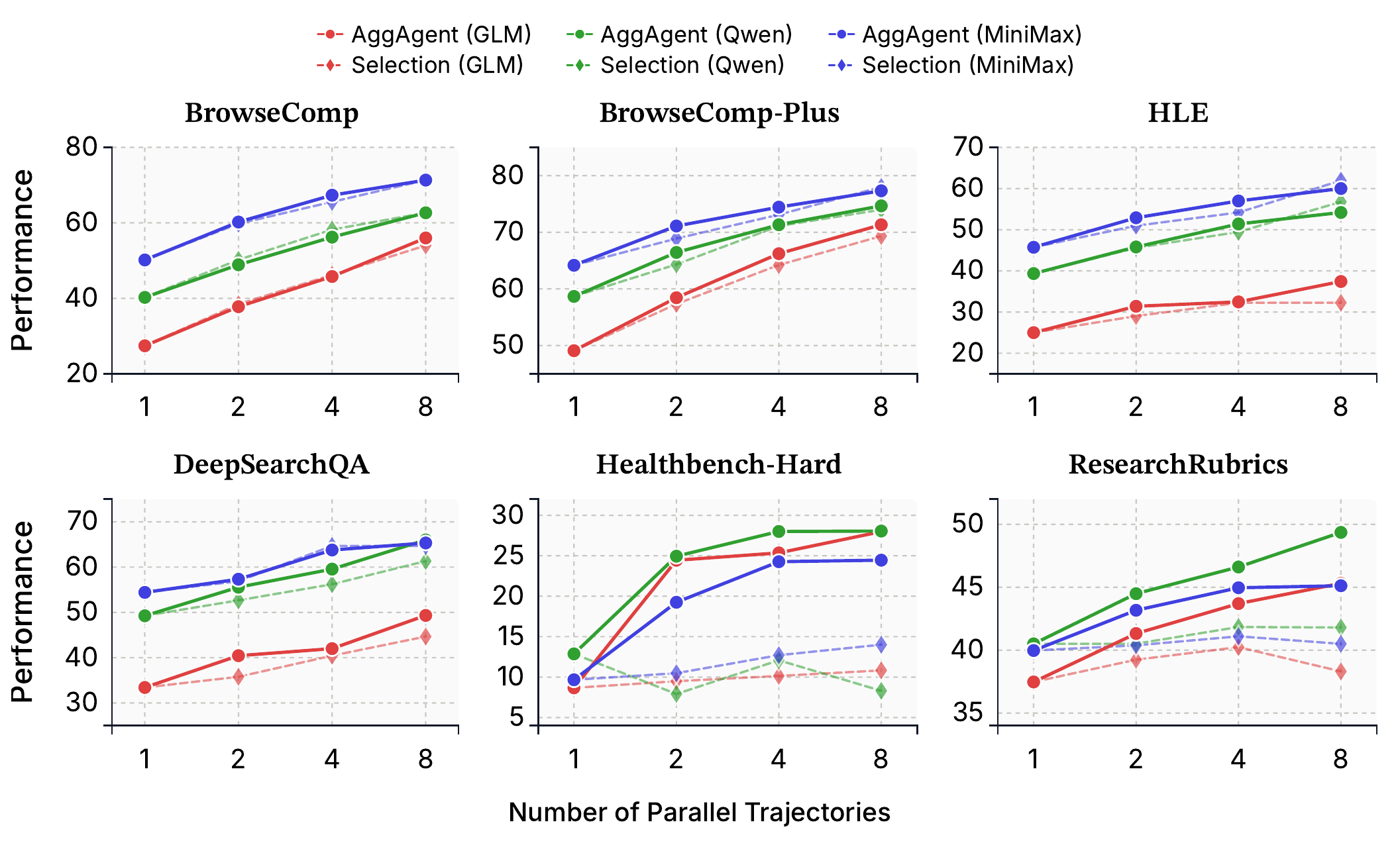}
    \caption{\textbf{Ablation of solution synthesis vs.\ best-trajectory selection}. \ours synthesizes a new solution from the collected trajectories; the selection variant selects the single best trajectory directly. Color indicates model (red: GLM-4.7-Flash, green: Qwen3.5-122B, blue: MiniMax-M2.5); line style and marker indicate method (solid + circle: \ours, dashed + diamond: selection variant).}
    \label{fig:bagg}
\end{figure}

\subsection{Output Design Ablation: Synthesis vs. Selection}
We ablate the output design of \ours by comparing it against a selection variant, which selects the single best trajectory and adopts its solution directly instead of synthesizing a new solution. As shown in Figure~\ref{fig:bagg}, \ours performs better overall, though the selection variant remains competitive in some settings. Notably, the selection variant performs poorly on deep research benchmarks, where \ours outperforms it by a large margin. We attribute this to a mismatch between selection and the nature of open-ended research tasks: quality is distributed across trajectories, so no single trajectory dominates, yet selection forces an all-or-nothing commitment. Synthesis mitigates this by making local, compositional judgments about which parts of each trajectory to incorporate, rather than committing to a global winner. This distinction is less pronounced for agentic search, where trajectories tend to be clearly correct or clearly wrong, making selection more straightforward. These results suggest that synthesis is often the preferable output design, consistent with our findings in Section~\ref{sec:stronger} and Appendix~\ref{sec:qual} that \ours can compose a correct solution from individually incorrect trajectories. We leave a more thorough ablation of output design to future work.

\subsection{Tool Design Ablations}
\label{sec:tool_design_ablation}

\begin{table}[t]
\centering
\footnotesize
\setlength{\tabcolsep}{3pt}
\captionsetup[subtable]{labelformat=simple,labelsep=space,font=bf}
\renewcommand{\thesubtable}{(\alph{subtable})}
\begin{subtable}[t]{0.55\linewidth}
\centering
\caption{Search strategy variants}
\label{tab:tool_design_ablation_search}
\vspace{0.25em}
\begin{tabular}{lccc}
\toprule
 & \textbf{ROUGE-L} & \textbf{BM25} & \textbf{Qwen3-Emb.} \\
\midrule
GLM-4.7-Flash & \textbf{47.90} & 47.37 & 45.41 \\
Qwen3.5-122B & \textbf{55.83} & 54.92 & 55.21 \\
MiniMax-M2.5  & 57.31 & 57.89 & \textbf{57.91} \\
\bottomrule
\end{tabular}
\end{subtable}\hfill
\begin{subtable}[t]{0.40\linewidth}
\centering
\caption{Disabling thinking traces}
\label{tab:tool_design_ablation_thinking}
\vspace{0.25em}
\begin{tabular}{lcc}
\toprule
 & \textbf{Full} & \textbf{No Think.} \\
\midrule
GLM-4.7-Flash & 47.90 & 45.13$_{\scriptscriptstyle (\textcolor{red}{-2.77})}$ \\
Qwen3.5-122B & 55.83 & 54.58$_{\scriptscriptstyle (\textcolor{red}{-1.25})}$ \\
MiniMax-M2.5  & 57.31 & 57.28$_{\scriptscriptstyle (\textcolor{red}{-0.03})}$ \\
\bottomrule
\end{tabular}
\end{subtable}

\caption{\textbf{Tool design ablations.} Scores are averaged across six benchmarks at $K{=}8$ and the same model is used for both rollout generation and aggregation. Table~\ref{tab:tool_design_ablation_search} compares retrieval backends for search\_trajectory: ROUGE-L, BM25, and Qwen3-Embedding-8B; bold indicates the best retrieval backend in each row. Table~\ref{tab:tool_design_ablation_thinking} compares the default full-trajectory setting (Full) with a variant that restricts access to internal thinking traces (No Think.); red subscripts show the score change relative to Full.}
\label{tab:tool_design_ablation}
\end{table}

\paragraph{Search strategy variants}
We ablate the retrieval backend used by the \texttt{search\_trajectory} tool by comparing ROUGE-L with BM25 and Qwen3-Embedding-8B. As shown in Table~\ref{tab:tool_design_ablation_search}, performance differences are small, and no retrieval strategy consistently dominates across model families. This suggests that \ours's gains are not driven by a particular search backend. We hypothesize that dense embedding retrieval is not uniformly advantageous for trajectory search because aggregation often requires locating exact strings, localized evidence, or tool-returned facts rather than broad semantic matches. The retrieval backend acts as a design tradeoff among retrieval quality, latency, and implementation cost, rather than as a determinant of \ours's effectiveness.

\paragraph{Disabling thinking traces}
The default \ours environment exposes full trajectories, including internal thinking traces, as our rollout agents generate complete execution logs. However, our method is also compatible with systems that do not return thinking traces. To isolate this factor, we evaluate a variant in which \ours cannot access internal reasoning text. As shown in Table~\ref{tab:tool_design_ablation_thinking}, removing thinking traces causes only a small drop in average performance, and the resulting variant still outperforms all the baselines in Table~\ref{tab:main_mini}. This indicates that \ours's gains primarily come from access to verifiable trajectory evidence rather than dependence on internal reasoning traces. Stronger models remain more robust, suggesting that they can aggregate effectively from tool observations alone.

\begin{figure}[t]
    \centering
    \includegraphics[width=\linewidth]{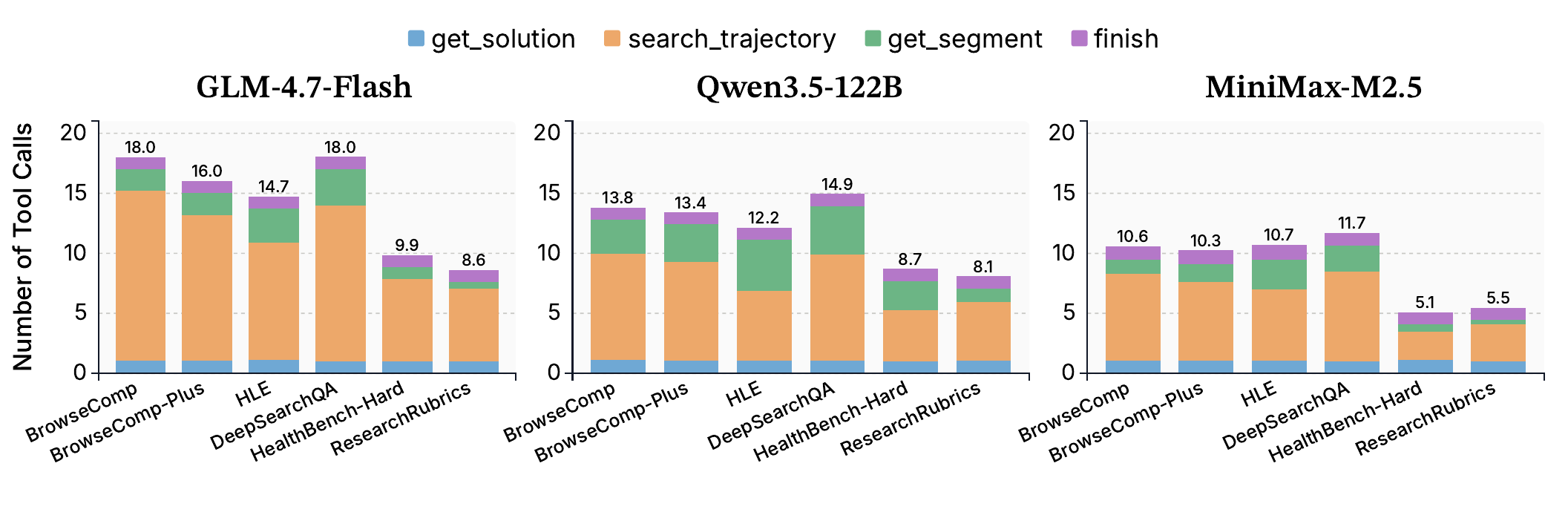}
    \caption{\textbf{Average number of tool calls per query by \ours.} Numbers above each bar indicate the average total tool calls per query. {search\_trajectory} dominates tool usage, while get\_solution and finish are each called approximately once per query. get\_segment is used more selectively, reflecting a coarse-to-fine strategy where \ours commits to full-content reads only when keyword-level search is insufficient.}
    \label{fig:tools}
\end{figure}

\subsection{Tool Usage Patterns}
In this section, we analyze the tool usage behaviour of \ours. As depicted in Figure~\ref{fig:tools}, \texttt{search\_trajectory} accounts for the majority of tool calls, while \texttt{get\_solution} and \texttt{finish} are each invoked approximately once per query, consistent with the intended workflow: solutions are surveyed once at the start and the agent terminates after the final synthesis. Meanwhile, \texttt{get\_segment} is used less often than \texttt{search\_trajectory}, suggesting that \ours reserves full-content reads for cases where keyword-level search alone is insufficient, diving deep only when actually needed. We also observe that stronger models tend to require fewer tool calls overall, suggesting more capable models reach a final solution more efficiently. Deep research tasks similarly exhibit fewer tool calls, potentially due to less need for resolving precise factual conflicts across trajectories.

\section{Related Work}
\paragraph{Long-horizon agents}
LLM agents have emerged as a powerful paradigm, extending language models beyond pure parametric generation to interactive problem solving in external environments. ReAct~\citep{yao2022react} has been the most prominent workflow, interleaving reasoning, action selection, and environment observations. This has enabled a wide range of applications, including agentic RAG~\citep{jin2025search, li2025search, jin2025empirical}, software engineering~\citep{jimenez2023swe, yang2024swe, wang2024openhands}, web navigation~\citep{yao2022webshop, chae2025web, gandhi2025go}, and deep research systems~\citep{OpenAIDeepResearch2025, AnthropicDeepResearch2025, HuggingfaceDeepResearch2025}. Recent work has sought to improve agentic capabilities through advances in RL algorithms~\citep{shao2024deepseekmath, shao2025dr, ritter2026llms}, agentic mid-training~\citep{su2025scaling}, and environment scaling~\citep{fang2025towards}. Our work is complementary to these efforts: we treat these agents as rollout models and study how to aggregate parallel trajectories into a single high-quality solution.

\paragraph{Test-time scaling}
Early work on test-time scaling for CoT tasks~\citep{kojima2022large, wei2022chain} such as math reasoning and coding showed clear gains from search-based strategies~\citep{yao2023tree}, extended thinking budgets~\citep{muennighoff2025s1, guo2025deepseek, snell2025scaling}, majority voting~\citep{wang2022self,brown2024large}, and learning-based aggregation~\citep{qi2025learning, zhao2025majority}.

However, test-time scaling for long-horizon agents is more challenging, as trajectories are lengthy, heterogeneous, and difficult to compare using simple surface-level signals (e.g., majority or confidence). Parallel scaling with LLM-based aggregation mitigates this problem by generating multiple independent trajectories and reasoning over them to produce a final solution. WebResearcher~\citep{qiao2025webresearcher} concatenates final answers and prompts an LLM to synthesize a solution, while ParallelMuse~\citep{li2025parallelmuse} compresses each trajectory into a summary report before aggregation, incurring high cost and information loss. Concurrent with our work, KARL~\citep{chang2026karl} feeds final answers back to the rollout agent with the same tools enabled, but it disregards full trajectories, and reusing rollout tools can be costly in practice. In contrast, \ours enables the aggregator to interact with trajectories in an agentic manner, supporting cross-trajectory reasoning without lossy compression or excessive cost. While \ours is training-free and effective with off-the-shelf LLMs, we leave fine-tuning the aggregator as a promising direction for future work.

Orthogonal to our focus on parallel scaling, several works have studied sequential scaling to extend the effective horizon of a single agent through context management~\citep{wu2025resum, yen2025lost, tang2025beyond, zeng2026glm}, context folding~\citep{sun2025scaling, ye2026agentfold}, or iterative refinement~\citep{zhu2026re, xiao2026meta}. These two directions could be combined in practice, for example, by using context management within each rollout and then applying \ours across the resulting trajectories (Appendix~\ref{app:seq_parallel}).

\section{Conclusion}

We introduce \ours, an aggregation agent for parallel scaling of long-horizon agentic tasks. By treating parallel trajectories as an interactive environment and navigating them via lightweight in-memory tools, \ours enables cross-trajectory reasoning at full fidelity, while avoiding both the information loss of compression and the prohibitive cost of loading all trajectories into context. Across six benchmarks and three models, \ours consistently outperforms all baselines, achieving Pareto-optimal performance and efficiency. Our findings establish agentic aggregation as a principled and cost-efficient paradigm for parallel test-time scaling.

\section*{Acknowledgments}
We thank Jeffrey Cheng, Simon Park, and all Princeton NLP members for their helpful discussions and feedback. This work is gratefully supported by an NSF CAREER award (IIS-2239290).

\bibliography{colm2026_conference}
\bibliographystyle{colm2026_conference}

\appendix
\section{Experimental Details}
\label{app:exp}
\subsection{Implementation}
\label{app:impl}

We serve all models via vLLM~\citep{kwon2023efficient} and use the Tongyi DeepResearch scaffold~\citep{team2025tongyi} for the agentic workflow. We enable interleaved reasoning and native function calling, with temperature~1.0, top-$p$~0.95, a 128K context window, 10K max output tokens, and at most 100 tool calls per rollout. Upon exhausting the context limit or tool call budget, the model is given one final opportunity to submit an answer without tool access. All experiments are conducted on 4$\times$H100 GPUs.

\subsection{Rollout Agent Tools}

\textbf{BrowseComp-Plus} The local corpus contains 100,195 documents. \texttt{search} uses Qwen3-Embedding-8B to retrieve the top-5 documents by score, with each snippet truncated to 512 tokens. \texttt{get\_document} returns the full text of a document by docid, truncated to 4,096 tokens following~\citet{su2025scaling}.
\begin{itemize}
[leftmargin=*,noitemsep,nolistsep,itemsep=3pt]
    \item \textbf{\texttt{search(query)}}: Perform a search on a knowledge source. Returns top-5 hits with docid, score, and snippet. The snippet contains the document's contents (may be truncated based on token limits).
    \item \textbf{\texttt{get\_document(docid)}}: Retrieve a full document by its docid.
\end{itemize}

\textbf{All other datasets} We adopt the tool implementation from SLIM~\citep{yen2025lost}. \texttt{search} queries Google Search via the Serper API~(\url{https://serper.dev}), returning the top-10 results with snippets of at most 150 characters. \texttt{visit} scrapes a URL via crawl4ai~(\url{https://github.com/unclecode/crawl4ai}) and extracts the snippet most relevant to a goal, ranked by ROUGE-L score~\citep{lin-2004-rouge}.

\begin{itemize}
[leftmargin=*,noitemsep,nolistsep,itemsep=3pt]
    \item \textbf{\texttt{search(query)}}: Performs a web search: supply a string `query'; the tool retrieves the top 10 results for the query.
    \item \textbf{\texttt{visit(url, goal)}}: Visit a webpage and return the relevant content based on the goal.
\end{itemize}

\subsection{Cost Calculation}
\label{app:pricing}
Let $g_{in}$ and $g_{out}$ denote the per-token input and output costs of a language model, and let $g_s$ and $g_v$ denote the per-call costs of \texttt{search} and \texttt{visit}, respectively. The total cost per question is $c = c_r + c_t + c_a$.

\textbf{Rollout cost} ($c_r$): Following~\citet{yen2025lost}, cached input tokens are excluded from the token count, as caching is standard in long-trajectory settings with shared context. The total rollout cost is $c_r = \sum_i \left( g_{in} \cdot \Delta t_{in}^{(i)} + g_{out} \cdot t_{out}^{(i)} \right)$, where $\Delta t_{in}^{(i)} = t_{in}^{(i)} - t_{in}^{(i-1)}$.

\textbf{Tool call cost} ($c_t$): For BrowseComp-Plus, which uses a local corpus, $c_t = 0$. For all other tasks, $c_t = g_s \cdot n_s + g_v \cdot n_v$, where $n_s$ and $n_v$ are the total numbers of \texttt{search} and \texttt{visit} calls.

\textbf{Aggregation cost} ($c_a$): Only LLM-based aggregation methods incur additional aggregation cost. For Solution Aggregation, $c_a$ is a single LLM call: $g_{in} \cdot t_{in} + g_{out} \cdot t_{out}$. For Summary Aggregation, this is summed over $K+1$ calls. For \ours, $c_a$ is computed identically to $c_r$, as it forms an agentic trajectory.

Cost is averaged over all instances for each benchmark. Pricing rates are given in Table~\ref{tab:pricing}. Prices are sourced from the respective provider websites: GLM-4.7-Flash (\url{https://docs.z.ai/guides/overview/pricing}), Qwen3.5-122B (\url{https://openrouter.ai/qwen/qwen3.5-122b-a10b}), MiniMax-M2.5 (\url{https://platform.minimax.io/docs/guides/pricing-paygo}), and tool calls (\url{https://www.firecrawl.dev/pricing}).

\begin{table}[h]
\centering
\begin{tabular}{lcc}
\toprule
 & \textbf{Input} & \textbf{Output} \\
\midrule
\multicolumn{3}{l}{\textit{Language Models (per 1M tokens)}} \\
\quad GLM-4.7-Flash & \$0.07 & \$0.40 \\
\quad Qwen-3.5-122B & \$0.26 & \$2.08 \\
\quad MiniMax-M2.5  & \$0.30 & \$1.20 \\
\midrule
\multicolumn{3}{l}{\textit{Tool Calls (per 1K queries)}} \\
\quad \texttt{search} & \multicolumn{2}{c}{\$0.50} \\
\quad \texttt{visit}  & \multicolumn{2}{c}{\$0.83} \\
\bottomrule
\end{tabular}
\caption{Pricing used for cost calculation.}
\label{tab:pricing}
\end{table}

\subsection{Latency Measurement}
We measure latency by processing one problem at a time, using 2$\times$H100 for GLM-4.7-Flash and 4$\times$H100 for Qwen3.5-122B and MiniMax-M2.5.

\textbf{Rollout latency} ($\ell_r$): Wall-clock time from the first step of the trajectory until termination.

\textbf{Aggregation latency} ($\ell_a$): For Solution Aggregation, the time of the single LLM call. For Summary Aggregation, we spawn $K$ summarization calls in parallel, wait for all to complete, then issue the final aggregation call; $\ell_a$ covers the full span. For \ours, $\ell_a$ is measured the same way as $\ell_r$.

For latency, we sample 30 instances and report the median for efficiency, as latency must be measured sequentially. All other experiments run with batched generation.

\subsection{Evaluation}
\label{app:eval}
We evaluate \text{Metric@}$K\in\{1,2,4,8\}$ over $N=8$ independent trajectories via bootstrapped sampling (Equation~\ref{eq:metric}), where $Q$ is the dataset, $S$ is a bootstrap sample of size $K$, and $\mathcal{M}$ is the scoring function between the predicted solution $\hat{y}$ and ground truth $y^*$. For LLM-based aggregation, evaluating all $\binom{N}{K}$ combinations is prohibitively expensive, so we cap the number of combinations to 3.

{\small
\begin{equation}
\text{Metric@}K = \frac{1}{|Q|}\sum_{q\in Q} \left( \frac{1}{\binom{N}{K}} \sum_{\substack{S \subseteq \{1,\dots,N\} \\ |S|=K}} \mathcal{M}(\hat{y}, y^*) \right)
\label{eq:metric}
\end{equation}
}

\subsection{Rollout Trajectory Statistics}
We provide the statistics of base rollout trajectories in Table~\ref{tab:base_stat}.
\begin{table}[h]
\centering
\resizebox{\linewidth}{!}{%
\begin{NiceTabular}{cl|w{c}{1cm}w{c}{1.8cm}w{c}{1.8cm}w{c}{1.8cm}w{l}{3.2cm}w{l}{1.8cm}}
\toprule
& \textbf{Task} & \textbf{\# Inst.} & \textbf{\# Turns} & \textbf{Tokens (K)} & \textbf{Output (K)} & \textbf{\# Tool Calls} & \textbf{Uniq. URLs} \\
\midrule\midrule

\Block[fill=red!8]{6-1}{\rotatebox{90}{\textbf{GLM-4.7-Flash}}}
& BrowseComp       & 150 & $71.4_{0.9}$  & $64.8_{1.0}$  & $0.64_{0.05}$  & $70.4_{0.9}\ (59.0/11.4)$   & $324.6_{9.4}$ \\
& BrowseComp-Plus  & 150 & $26.0_{0.8}$  & $72.0_{1.8}$  & $0.53_{0.01}$  & $25.1_{0.8}\ (23.1/2.0)$    & $50.1_{0.7}$  \\
& HLE              & 155 & $34.4_{1.7}$  & $40.9_{2.2}$  & $1.39_{0.12}$  & $31.9_{1.5}\ (22.8/9.1)$    & $86.8_{5.1}$  \\
& DeepSearchQA     & 150 & $53.4_{2.4}$  & $48.3_{1.5}$  & $0.71_{0.09}$  & $52.6_{2.4}\ (37.9/14.7)$   & $126.8_{3.9}$ \\
& Healthbench-Hard & 100 & $17.8_{1.6}$  & $27.0_{2.5}$  & $1.74_{0.07}$  & $16.9_{1.6}\ (7.9/9.0)$     & $52.5_{6.1}$  \\
& ResearchRubrics  & 101 & $34.0_{1.8}$  & $48.2_{2.1}$  & $4.23_{0.11}$  & $33.1_{1.8}\ (15.9/17.2)$   & $121.4_{6.7}$ \\

\midrule

\Block[fill=customgreen]{6-1}{\rotatebox{90}{\textbf{Qwen3.5-122B}}}
& BrowseComp       & 150 & $59.4_{1.0}$  & $61.7_{1.8}$  & $1.36_{0.19}$  & $58.7_{1.0}\ (52.2/6.5)$    & $349.2_{9.4}$ \\
& BrowseComp-Plus  & 150 & $21.2_{0.3}$  & $60.2_{1.0}$  & $1.23_{0.11}$  & $20.4_{0.4}\ (18.8/1.7)$    & $38.7_{0.6}$  \\
& HLE              & 155 & $26.5_{0.8}$  & $35.6_{1.0}$  & $1.86_{0.10}$  & $25.2_{0.8}\ (18.2/7.0)$    & $84.2_{2.8}$  \\
& DeepSearchQA     & 150 & $40.4_{1.4}$  & $45.9_{1.6}$  & $1.04_{0.04}$  & $39.4_{1.4}\ (26.9/12.6)$   & $116.5_{3.9}$ \\
& Healthbench-Hard & 100 & $\phantom{0}7.4_{0.2}$   & $12.1_{0.3}$  & $1.71_{0.03}$  & $\phantom{0}6.4_{0.2}\ (2.3/4.1)$      & $20.2_{1.0}$  \\
& ResearchRubrics  & 101 & $15.1_{0.0}$  & $24.0_{0.0}$  & $4.66_{0.00}$  & $14.1_{0.0}\ (7.4/6.6)$     & $65.6_{0.0}$  \\

\midrule

\Block[fill=customblue]{6-1}{\rotatebox{90}{\textbf{MiniMax-M2.5}}}
& BrowseComp       & 150 & $54.1_{1.2}$  & $50.1_{1.4}$  & $1.41_{0.10}$  & $53.0_{1.2}\ (42.1/10.9)$   & $189.5_{5.8}$ \\
& BrowseComp-Plus  & 150 & $23.7_{0.5}$  & $64.8_{1.3}$  & $1.11_{0.12}$  & $22.8_{0.5}\ (21.4/1.3)$    & $48.6_{1.2}$  \\
& HLE              & 155 & $28.4_{1.0}$  & $42.3_{1.7}$  & $2.08_{0.19}$  & $26.8_{1.1}\ (17.9/8.9)$    & $79.4_{5.3}$  \\
& DeepSearchQA     & 150 & $44.0_{0.7}$  & $49.1_{1.1}$  & $1.33_{0.06}$  & $42.9_{0.7}\ (26.6/16.3)$   & $122.0_{2.3}$ \\
& Healthbench-Hard & 100 & $\phantom{0}8.7_{0.4}$   & $13.2_{0.6}$  & $1.68_{0.08}$  & $\phantom{0}7.6_{0.4}\ (3.4/4.3)$      & $27.6_{1.6}$  \\
& ResearchRubrics  & 101 & $20.9_{0.6}$  & $27.7_{0.5}$  & $3.89_{0.07}$  & $19.9_{0.6}\ (12.2/7.7)$    & $101.8_{5.1}$ \\

\bottomrule
\end{NiceTabular}%
}
\caption{Trajectory statistics of base rollouts (8 per problem) across three models and six benchmarks. \textbf{\# Inst.}: number of evaluation instances. \textbf{\# Turns}: number of agent turns per trajectory. \textbf{Tokens (K)}: total input+output tokens in thousands. \textbf{Output (K)}: output tokens in thousands. \textbf{\# Tool Calls}: total tool calls, broken down into search and visit calls (shown in parentheses; {get\_document} is used instead of {visit} for BrowseComp-Plus). \textbf{Uniq.\ URLs}: number of unique URLs visited. All values are averaged over problems and rollouts, and subscripts denote standard deviation across rollout sets.}
\label{tab:base_stat}
\end{table}

\section{Prompts}
\label{app:prompts}
\paragraph{Rollout agent}
We use the system prompt from Tongyi DeepResearch~\citep{team2025tongyi} (Figure~\ref{fig:rollout_qa}), with an additional instruction from Dr.\ Tulu~\citep{shao2025dr} appended for deep research tasks (Figure~\ref{fig:rollout_dr}). For agentic search tasks, we format the user message following BrowseComp~\citep{wei2025browsecomp} (Figure~\ref{fig:rollout_user}).

\paragraph{LLM-as-a-Judge}
For BrowseComp, BrowseComp-Plus, and HLE, we find that the original evaluation prompt occasionally produces false judgments, hence we instead use the prompt from~\citet{zhu2026re}. For ResearchRubrics, we find that the original prompt is unreliable for negative rubrics --- consistent with~\citet{hu2025step} --- so we append a clarifying section to the original prompt (Figure~\ref{fig:eval_rr}). For DeepSearchQA and Healthbench-Hard, we use the official evaluation prompts.

\paragraph{LLM-based aggregation}
\label{app:prompts_aggagent}
For Solution Aggregation and Summary Aggregation, we use the prompt from~\citet{li2025parallelmuse}. For \ours, we present the prompts in Figure~\ref{fig:aggagent_prompt_qa} (agentic search) and Figure~\ref{fig:aggagent_prompt_dr} (deep research), and the full tool descriptions in Figure~\ref{fig:aggagent_tools}. We vary the solution field description of the \texttt{finish} tool (Figure~\ref{fig:aggagent_tools_ab}) in the following cases: 1) Deep research tasks: the model is instructed to generate a long-form report. 2) Qwen3.5-122B: instructing the model to output in XML format sometimes causes tool parsing errors, so we use plain text format instead. This applies only to agentic search tasks.

\section{Results}

\subsection{Main Results}
\label{app:main_results}
We present the full results for Section~\ref{sec:main_results} in Tables~\ref{tab:main_glm},~\ref{tab:main_qwen}, and~\ref{tab:main_minimax}. \ours achieves competitive performance not only at $K=8$ but also at $K=2$ and $K=4$.

\subsection{Cost and Latency vs. Performance}
\label{app:cost_results}
\begin{figure}[t]
    \centering
    \includegraphics[width=\linewidth]{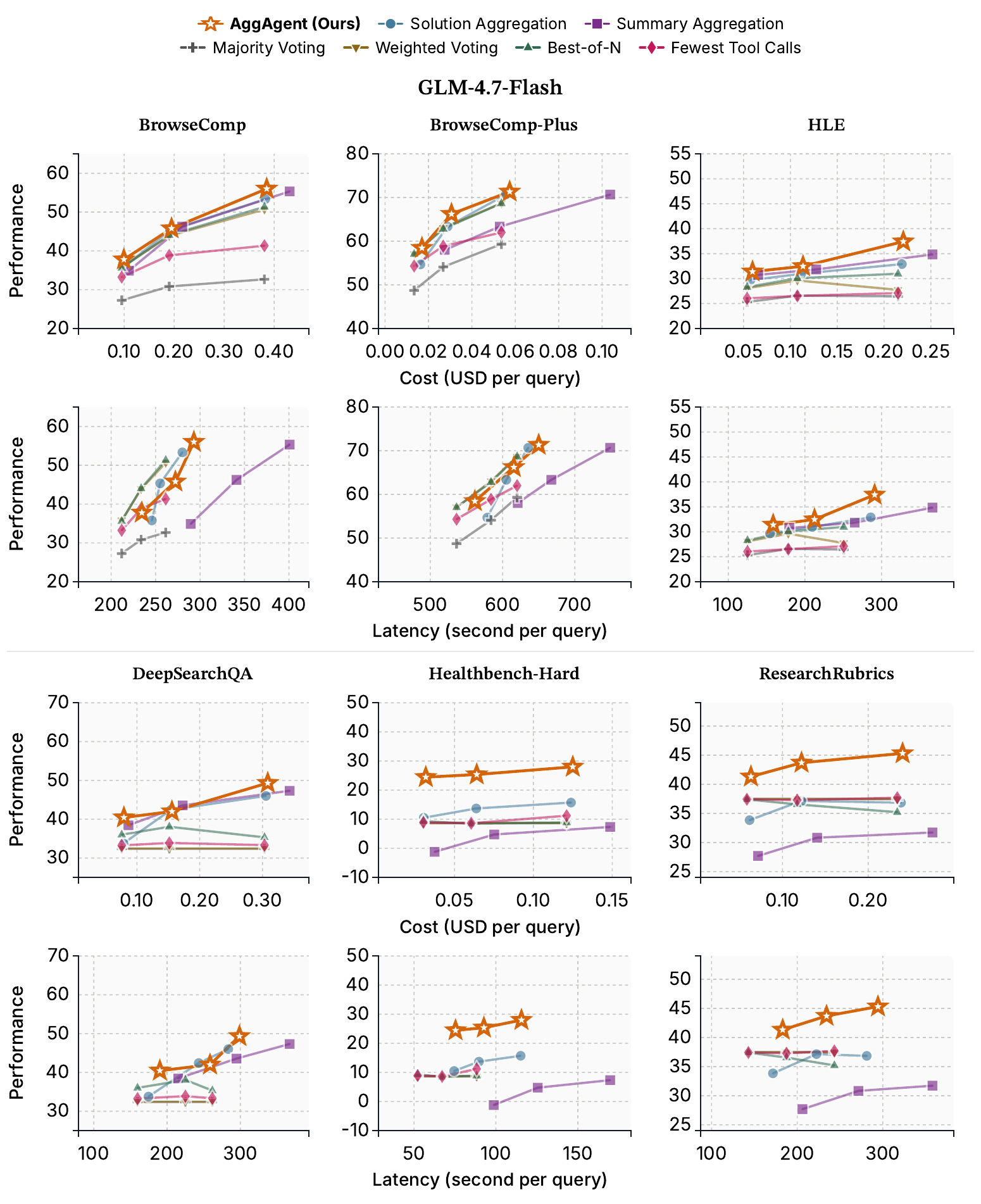}
    \caption{Performance-efficiency tradeoff of aggregation methods across six benchmarks at varying numbers of parallel samples $K=\{2,4,8\}$ using GLM-4.7-Flash as both the rollout agent and the aggregator. For each group of three benchmarks, the first row shows performance vs.\ cost (USD per query) and the second row shows performance vs.\ latency (seconds per query). \ours consistently achieves higher performance at comparable or lower cost and latency than existing aggregation strategies.}
    \label{fig:grouped_glm}
\end{figure}

\begin{figure}[t]
    \centering
    \includegraphics[width=\linewidth]{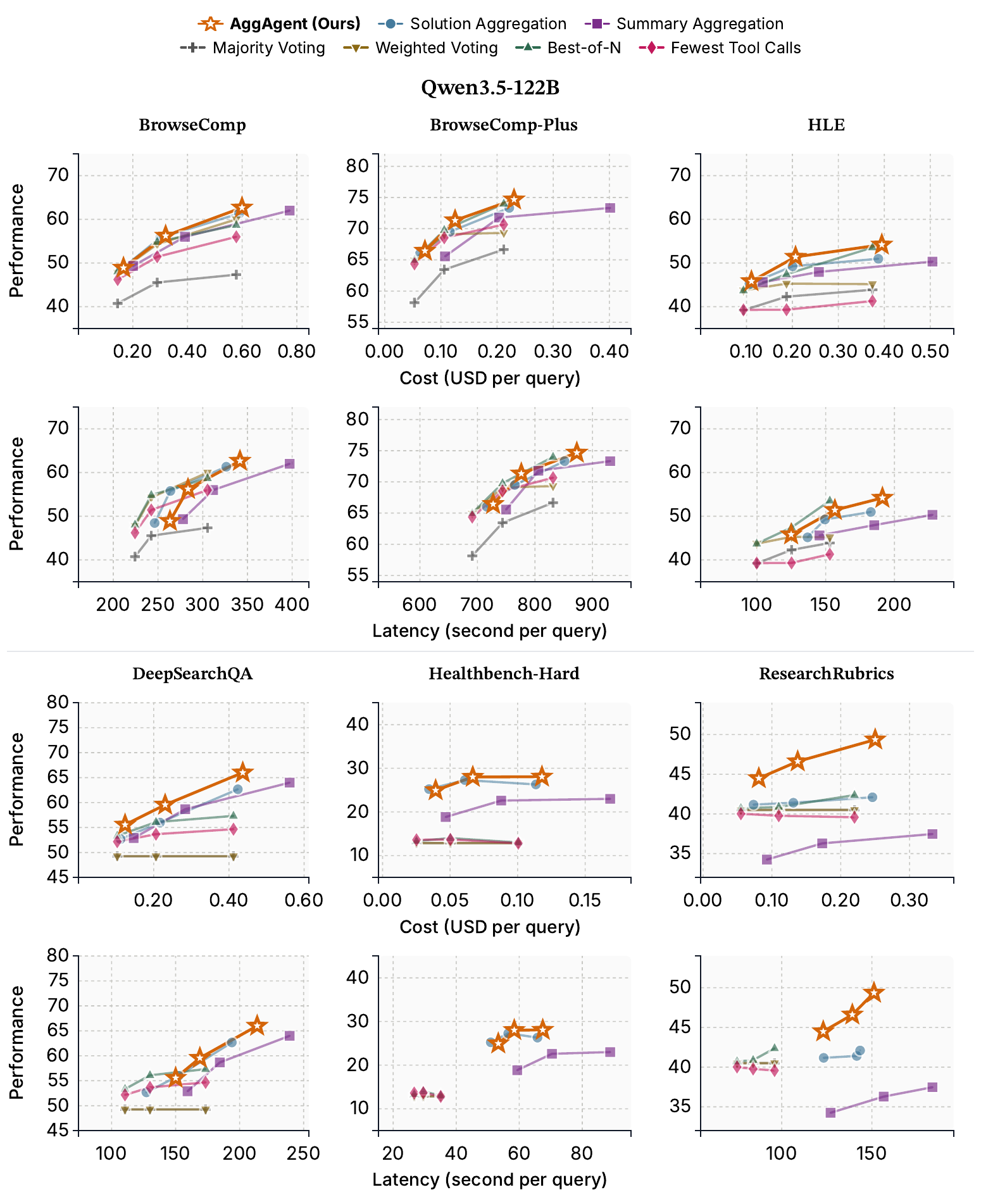}
    \caption{Performance-efficiency trade-off of aggregation methods using Qwen3.5-122B as both the rollout agent and the aggregator. The figure is formatted the same as Figure~\ref{fig:grouped_glm}.}
    \label{fig:grouped_qwen}
\end{figure}

\begin{figure}[t]
    \centering
    \includegraphics[width=\linewidth]{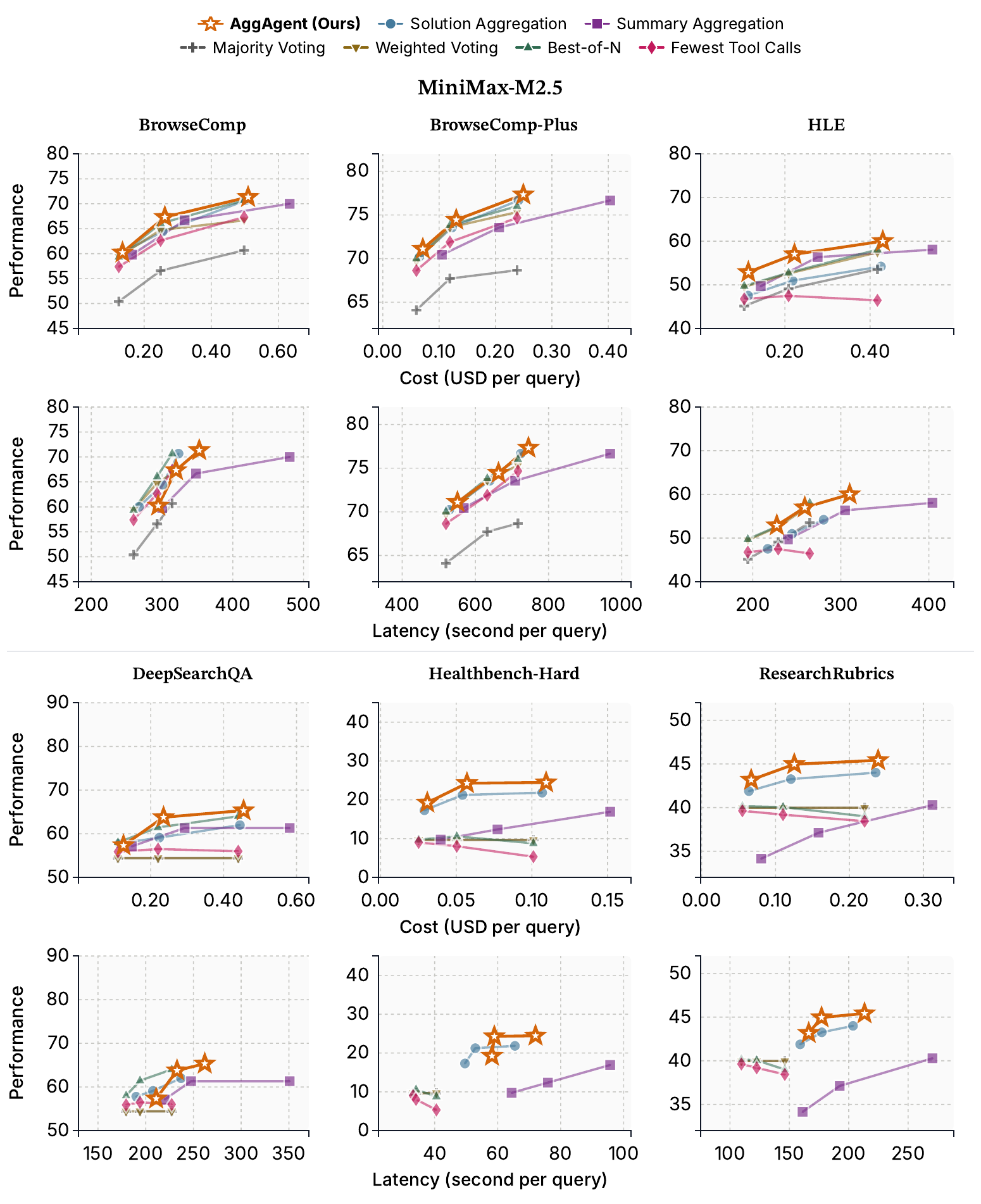}
    \caption{Performance-efficiency trade-off of aggregation methods using MiniMax-M2.5 as both the rollout agent and the aggregator. The figure is formatted the same as Figure~\ref{fig:grouped_glm}.}
    \label{fig:grouped_minimax}
\end{figure}

We present the full results for Section~\ref{sec:cost} in Figures~\ref{fig:grouped_glm},~\ref{fig:grouped_qwen}, and~\ref{fig:grouped_minimax}. The per-benchmark results are consistent with the averaged figures in the main paper, with \ours achieving optimal performance relative to cost and latency.

We also present the cost and latency breakdown in Tables~\ref{tab:cost_latency_glm},~\ref{tab:cost_latency_qwen}, and~\ref{tab:cost_latency_minimax}. Across all settings, \ours consistently incurs aggregation cost much lower than Summary Aggregation and comparable to Solution Aggregation.

\subsection{Stronger Models for Aggregation}
We present the full results for Section~\ref{sec:stronger} in Table~\ref{tab:minimax}. While a stronger aggregator (MiniMax) generally improves performance across all LLM-based aggregation methods, \ours remains the most effective.

\subsection{Confidence Calibration}
\label{sec:calib}
\begin{figure}[h]
    \centering
    \includegraphics[width=\linewidth]{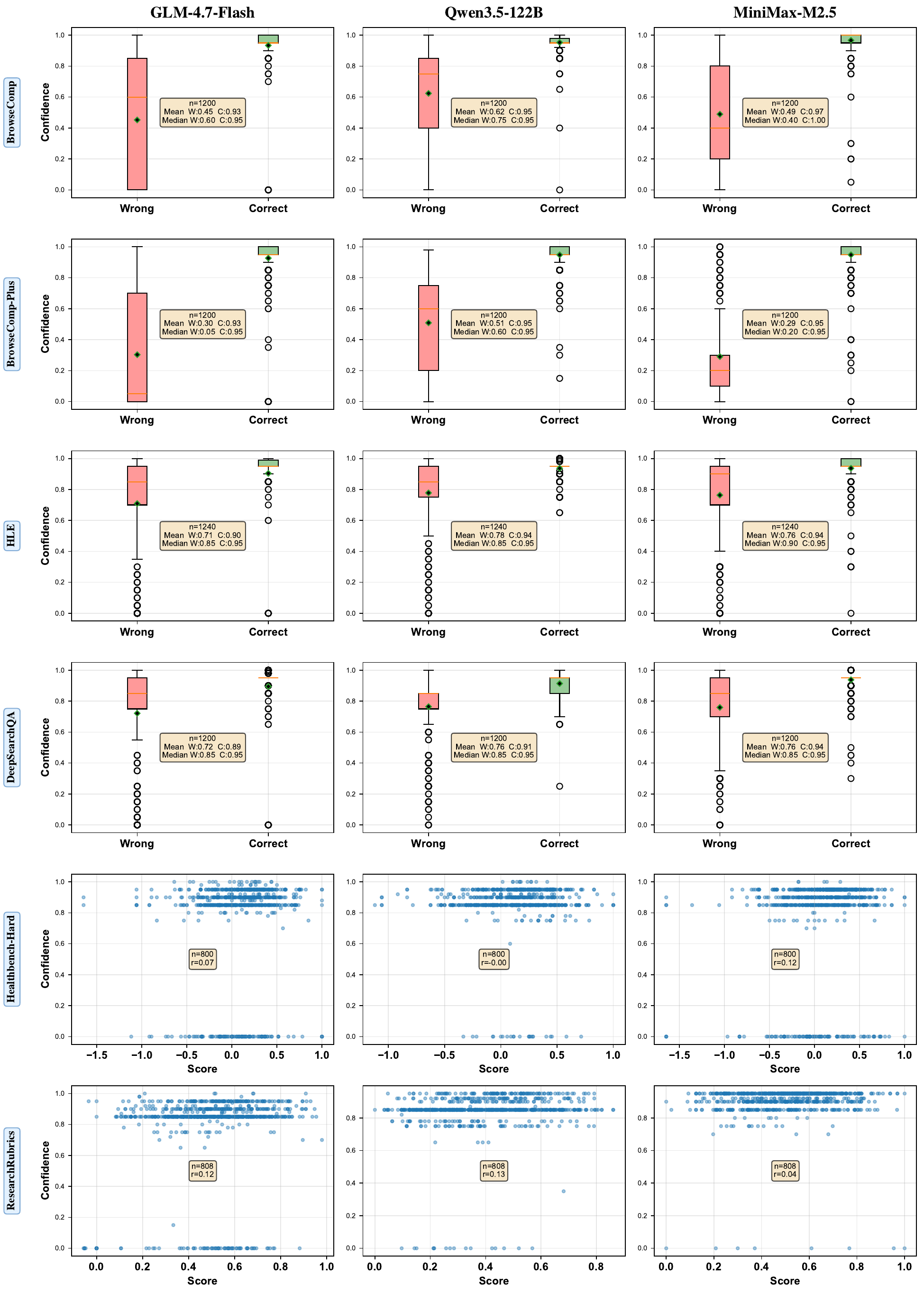}
    \caption{{Confidence calibration across all six benchmarks.} Each row is a benchmark and each column is a model. For binary-correctness (agentic search) benchmarks, box plots show self-reported confidence by outcome (\textbf{W}rong vs.\ \textbf{C}orrect). For continuous-score (deep research) benchmarks, scatter plots show confidence vs.\ rubric score with Pearson $r$. Models are well-calibrated on BrowseComp-Plus and BrowseComp, moderately calibrated on HLE and DeepSearchQA, but poorly calibrated on Healthbench-Hard and ResearchRubrics ($r \approx 0$).}
    \label{fig:confidence_grid}
\end{figure}

While prior work has demonstrated the effectiveness of confidence-based aggregation strategies such as Weighted Majority Voting and Best-of-N~\citep{wei2025browsecomp, li2025parallelmuse}, we find that these methods do not consistently generalize across benchmarks. As illustrated in Figure~\ref{fig:confidence_grid}, all three models show well-calibrated confidence on BrowseComp-Plus and BrowseComp (large $C - W$ gap), but are less calibrated on HLE and DeepSearchQA. Moreover, on Healthbench-Hard and ResearchRubrics, confidence is essentially uncorrelated with quality ($r \approx 0$). We hypothesize that this miscalibration explains why confidence-based methods plateau or even degrade as $K$ increases on those tasks. In contrast, LLM-based aggregation, and especially \ours, scales reliably with $K$ across all benchmarks.

\section{Analysis and Discussion}

\subsection{Trajectory-RAG Baseline}
\label{app:rag_baseline}

\ours retrieves evidence from trajectory logs on demand, which raises a natural question: how much of its benefit comes from adaptive, agentic retrieval rather than retrieval over trajectories itself? To study this, we compare against a non-agentic retrieval-augmented generation~\citep{lewis2020retrieval} pipeline that has access to the same trajectory logs but performs retrieval only once before synthesis.

Trajectory-RAG baseline indexes trajectory chunks, retrieves relevant evidence, and feeds the retrieved chunks to a standard LLM aggregator without iterative tool use. Conceptually, this baseline can be viewed as \SolAgg augmented with retrieved trajectory evidence. Unlike \ours, it does not interactively inspect candidate answers, follow up on disagreements, or issue additional retrieval actions based on intermediate hypotheses. 

We evaluate three retrieval backends: ROUGE-L, BM25, and Qwen3-Embedding-8B. In preliminary experiments, we found that filling the full context window with retrieved chunks performs better than using a fixed $k$ such as 25 or 50, improving performance by roughly 2 points. We also include the final solution candidates, which provide an additional 1--2 point gain.

\begin{table}[H]
\centering
\small
\setlength{\tabcolsep}{4pt}
\resizebox{\linewidth}{!}{%
\begin{tabular}{lcccccc}
\toprule
 & \textbf{\SolAgg} & \textbf{\SummAgg} & \textbf{RAG (ROUGE-L)} & \textbf{RAG (BM25)} & \textbf{RAG (Qwen3)} & \textbf{\ours} \\
\midrule
GLM-4.7-Flash & 42.58 & 41.21 & 44.10 & 44.45 & 43.66 & \textbf{47.90} \\
Qwen3.5-122B & 52.78 & 51.69 & 51.27 & 51.91 & 52.07 & \textbf{55.83} \\
MiniMax-M2.5  & 54.90 & 53.88 & 56.32 & 55.70 & 55.24 & \textbf{57.31} \\
\bottomrule
\end{tabular}%
}
\caption{\textbf{Trajectory-RAG baseline.} Scores are averaged across six benchmarks at $K{=}8$, with the same model used for both rollout generation and aggregation. RAG variants index trajectory chunks, retrieve evidence with ROUGE-L, BM25, or Qwen3-Embedding-8B, and feed the retrieved chunks plus final solution candidates to a non-agentic LLM aggregator.}
\label{tab:rag_baseline}
\end{table}

As shown in Table~\ref{tab:rag_baseline}, trajectory-RAG is competitive, confirming the value of retrieving evidence from trajectories. However, \ours consistently performs better across model families, suggesting that one-shot retrieval is not sufficient. \ours benefits from adaptively inspecting candidate answers, following up on disagreements, and iteratively verifying evidence before synthesis.

\subsection{Combining Sequential and Parallel Scaling}
\label{app:seq_parallel}

Our main experiments study aggregation over independently generated rollouts without additional context management. We further test whether \ours remains effective when the base rollout agent is strengthened with a simple sequential scaling technique. Specifically, we evaluate Qwen3.5-122B on BrowseComp with a discard-all policy at 80\% of the context window following~\citet{liu2025deepseek}. In the original rollouts, around 15\% of trajectories exceeded the context limit, and another 15\% exceeded the tool-call budget. Applying context management improves Pass@1 from 40.25 to 46.50.

We then apply aggregation on top of these stronger rollouts. For \SummAgg, trajectories exceeding the context window are handled with cascading summarization over chunks, so the full trajectory is compressed rather than simply truncated.

\begin{table}[H]
\centering
\footnotesize
\setlength{\tabcolsep}{3.8pt}
\resizebox{0.82\linewidth}{!}{%
\begin{tabular}{lrccccccc}
\toprule
\multirow{2}{*}{$\mathbf K$} & \multirow{2}{*}{\textbf{Pass}} & \multicolumn{4}{c}{\textbf{Heuristic Aggregation}} & \multicolumn{3}{c}{\textbf{LLM-based Aggregation}} \\
\cmidrule(lr){3-6}\cmidrule(lr){7-9}
 & & \textbf{\MV} & \textbf{\WMV} & \textbf{\BoN} & \textbf{\FewTool} & \textbf{\SolAgg} & \textbf{\SummAgg} & \textbf{\ours} \\
\midrule \midrule
1 & 46.50 & - & - & - & - & - & - & - \\
2 & 56.07 & 46.31 & 53.45 & 53.43 & 53.43 & 55.11 & \textbf{56.00} & \textbf{56.00} \\
4 & 63.90 & 51.85 & 58.40 & 58.60 & 58.60 & 59.78 & 60.44 & \textbf{61.33} \\
8 & 70.67 & 54.00 & 58.67 & 62.67 & 55.33 & 65.33 & 66.67 & \textbf{68.00} \\
\bottomrule
\end{tabular}%
}
\caption{\textbf{Combining sequential and parallel scaling.} Results are on BrowseComp using Qwen3.5-122B with a simple context management policy for base rollouts. Best aggregation result per row is bolded.}
\label{tab:seq_parallel}
\vspace{-8pt}
\end{table}

As shown in Table~\ref{tab:seq_parallel}, \ours demonstrates consistent gains over a wide range of Pass@1 values. Therefore, context management and agentic aggregation are complementary. Improving the base rollout raises absolute performance, while \ours provides additional gains over the aggregation baselines. 

\subsection{Failure Mode Analysis}
\label{app:failure_modes}

In this section, we analyze cases where Pass@$K$ succeeds (i.e., at least one trajectory produces a correct answer) but \ours fails. We identify four failure modes: (1) \textbf{Retrieval failure}, where the trajectory segment containing decisive evidence is not retrieved; (2) \textbf{Evidence interpretation failure}, where correct information is retrieved but the aggregator misinterprets or ignores it; (3) \textbf{Conflict resolution failure}, where the aggregator is misled by plausible but incorrect trajectories; and (4) \textbf{Synthesis failure}, which occurs especially on deep research tasks when correct partial evidence is not integrated into a coherent final response.

We manually annotate failure cases from MiniMax-M2.5 and report the breakdown in Table~\ref{tab:failure_modes}. The dominant source of failure is evidence interpretation, suggesting that improving the aggregator's ability to verify and reason over retrieved evidence is an important direction for closing the gap to Pass@$K$.

\begin{table}[H]
\centering
\small
\setlength{\tabcolsep}{5pt}
\resizebox{0.82\linewidth}{!}{%
\begin{tabular}{lrrrrr}
\toprule
\textbf{Dataset} & $N$ & \textbf{Retrieval} & \textbf{Evidence Interpretation} & \textbf{Conflict Resolution} & \textbf{Synthesis} \\
\midrule
BrowseComp      & 3  & 0.0\%  & 66.7\% & 33.3\% & 0.0\% \\
BrowseComp-Plus & 2  & 0.0\%  & 0.0\%  & 0.0\%  & 100.0\% \\
HLE             & 19 & 10.5\% & 68.4\% & 10.5\% & 10.5\% \\
DeepSearchQA    & 45 & 11.1\% & 57.7\% & 11.5\% & 19.2\% \\
\bottomrule
\end{tabular}%
}
\caption{\textbf{Failure mode analysis.} We manually annotate MiniMax-M2.5 cases where Pass@$8$ succeeds but \ours fails. $N$ denotes the number of annotated failure cases.}
\label{tab:failure_modes}
\vspace{-8pt}
\end{table}

\vspace{-1.0em}

\subsection{Qualitative Analysis}
\label{sec:qual}

\begin{figure}[tbp]
    \centering
    \includegraphics[width=\linewidth]{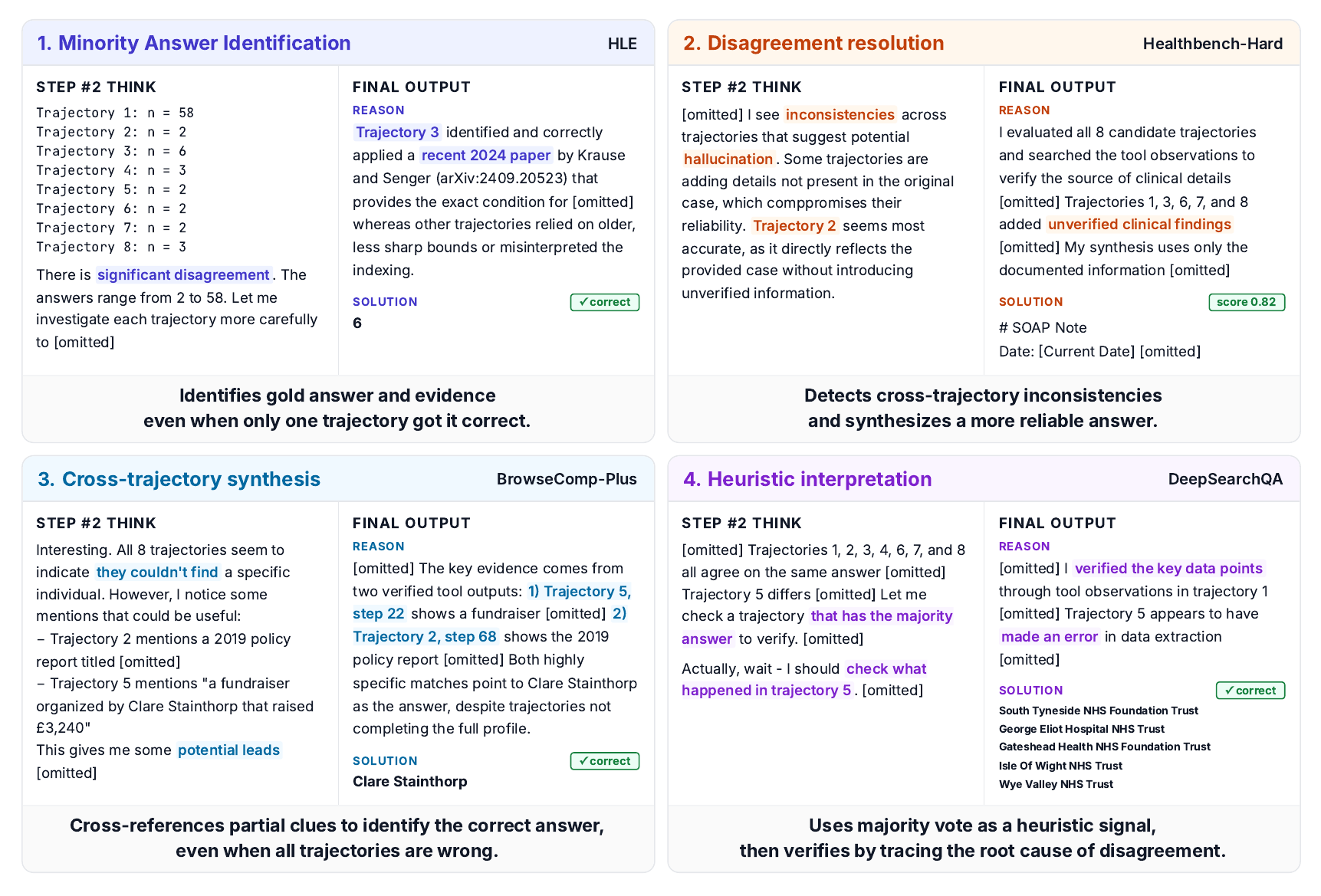}
    \caption{{Qualitative examples illustrating four key behaviours of \ours}.}
    \label{fig:qual}
\end{figure}

We qualitatively analyze cases where \ours succeeds in producing correct answers. \ours demonstrates four key behaviours (Figure~\ref{fig:qual}): (1) \textbf{Minority answer identification}: identifying the correct answer from multiple trajectories, even when it is held by only a minority, (2) \textbf{Disagreement resolution}: resolving inconsistencies and conflicts across trajectories to arrive at a coherent answer, (3) \textbf{Cross-trajectory synthesis}: constructing a correct solution by reasoning over trajectories that are all incorrect, and (4) \textbf{Heuristic interpretation}: leveraging heuristic signals such as majority vote and confidence scores to guide aggregation, while not relying on them entirely. Together, these behaviours highlight \ours's capacity for nuanced, evidence-driven reasoning across parallel trajectories beyond simple voting or selection.

\clearpage

\begin{table}[htbp]
\centering
\vspace{30pt}
\footnotesize
\resizebox{\linewidth}{!}{%
\setlength{\tabcolsep}{3.8pt}
\begin{tabular}{lrcccccccc}
\toprule
\multirow{2}{*}{\textbf{Task}} & \multirow{2}{*}{$\mathbf K$} & \multirow{2}{*}{\textbf{Pass}} & \multicolumn{4}{c}{\textbf{Heuristic Aggregation}} & \multicolumn{3}{c}{\textbf{LLM-based Aggregation}} \\
\cmidrule(lr){4-7}\cmidrule(lr){8-10}
 & & & \textbf{\MV} & \textbf{\WMV} & \textbf{\BoN} & \textbf{\FewTool} & \textbf{\SolAgg} & \textbf{\SummAgg} & \textbf{\ours} \\
\midrule \midrule

\multirow{4}{*}{BrowseComp}
& 1 & $27.42_{2.17}$ & - & - & - & - & - & - & -  \\
& 2 & $37.86_{2.63}$ & $27.26_{2.36}$ & $35.79_{3.03}$ & $35.60_{3.04}$ & $33.24_{2.47}$ & $35.78_{1.13}$ & $34.89_{2.45}$ & $\mathbf{37.78}_{1.37}$ \\
& 4 & $49.21_{2.04}$ & $30.83_{1.82}$ & $43.87_{2.41}$ & $44.06_{2.64}$ & $38.84_{1.93}$ & $45.33_{0.0}$ & $\mathbf{46.22}_{0.83}$ & $45.78_{1.13}$ \\
& 8 & $58.67$ & $32.67$ & $50.67$ & $51.33$ & $41.33$ & $53.33$ & $55.33$ & $\mathbf{56.00}$ \\

\midrule

\multirow{4}{*}{BrowseComp-Plus}
& 1 & $49.08_{3.09}$ & - & - & - & - & - & - & - \\
& 2 & $59.64_{1.76}$ & $48.71_{3.03}$ & $57.02_{2.33}$ & $57.05_{2.08}$ & $54.31_{2.41}$ & $54.67_{3.40}$ & $58.00_{1.44}$ & $\mathbf{58.44}_{1.37}$ \\
& 4 & $67.36_{1.54}$ & $54.10_{1.98}$ & $63.02_{2.05}$ & $62.81_{1.92}$ & $58.86_{1.81}$ & $63.33_{1.89}$ & $63.33_{0.54}$ & $\mathbf{66.22}_{0.83}$ \\
& 8 & $72.00$ & $59.33$ & $68.67$ & $68.67$ & $62.00$ & $70.67$ & $70.67$ & $\mathbf{71.33}$ \\

\midrule

\multirow{4}{*}{HLE}
& 1 & $25.00_{2.36}$ & - & - & - & - & - & - & -  \\
& 2 & $34.01_{2.38}$ & $25.32_{2.15}$ & $28.09_{2.39}$ & $28.29_{2.57}$ & $26.04_{2.35}$ & $29.68_{2.30}$ & $30.75_{0.80}$ & $\mathbf{31.40_{2.49}}$ \\
& 4 & $42.88_{1.99}$ & $26.58_{1.68}$ & $29.62_{1.49}$ & $30.08_{1.78}$ & $26.53_{2.08}$ & $30.97_{1.05}$ & $31.83_{1.61}$ & $\mathbf{32.47_{0.61}}$ \\
& 8 & $50.97$ & $26.45$ & $27.74$ & $30.97$ & $27.10$ & {$32.90$} & {$34.84$} & $\mathbf{37.42}$ \\

\midrule

\multirow{4}{*}{DeepSearchQA}
& 1 & $32.42_{4.36}$ & - & - & - & - & - & - & -\\
& 2 & $43.55_{4.01}$ & - & - & $36.02_{3.41}$ & $33.26_{3.90}$ & $33.78_{2.27}$ & $38.44_{2.06}$ & $\mathbf{40.44}_{1.57}$  \\
& 4 & $54.26_{2.90}$ & - & - & $38.06_{2.34}$ & $33.89_{2.20}$ & $42.44_{1.26}$ & $\mathbf{43.56}_{1.75}$ & ${42.00}_{1.44}$ \\
& 8 & $64.00$ & - & - & $35.33$ & $33.33$ & $46.00$ & $47.33$ & $\mathbf{49.33}$ \\

\midrule

\multirow{4}{*}{Healthbench-Hard}
& 1 & $8.67_{1.98}$ & - & - & - & - & - & - & -  \\
& 2 & $21.34_{1.65}$ & - & - & $9.80_{2.15}$ & $9.32_{1.72}$ & $10.50_{2.10}$ & $-1.21_{0.46}$ & $\mathbf{24.45}_{0.52}$ \\
& 4 & $31.96_{1.37}$ & - & - & $9.65_{1.93}$ & $8.48_{1.44}$ & $13.70_{2.18}$ & $4.76_{1.39}$ & $\mathbf{25.36_{1.59}}$ \\
& 8 & $40.91$ & - & - & $9.91$ & $8.90$ & $15.72$ & $7.35$ & $\mathbf{27.99}$ \\

\midrule

\multirow{4}{*}{ResearchRubrics}
& 1 & $37.47_{0.97}$ & - & - & - & - & - & - & -  \\
& 2 & $42.71_{0.53}$ & - & - & $37.46_{0.93}$ & $37.41_{0.78}$ & $33.84_{0.45}$ & $27.69_{0.84}$ & $\mathbf{41.32}_{0.72}$ \\
& 4 & $46.57_{0.30}$ & - & - & $37.68_{0.78}$ & $36.55_{0.74}$ & $37.15_{0.46}$ & $30.83_{0.37}$ & $\mathbf{43.70}_{0.58}$ \\
& 8 & $49.79$ & - & - & $37.70$ & $35.21$ & $36.84$ & $31.72$ & $\mathbf{45.31}$ \\

\midrule \midrule

\multirow{4}{*}{\shortstack{\textbf{Average}\\\textbf{(GLM-4.7-Flash)}}}
& 1 & $30.01$ & - & - & - & - & - & - & -  \\
& 2 & $39.85$ & $29.98$ & $33.24$ & $34.04$ & $32.26$ & 33.04 & 31.43 & \textbf{38.97} \\
& 4 & $48.71$ & $31.68$ & $35.85$ & $37.06$ & $33.86$ & 38.82 & 36.76 & \textbf{42.59} \\
& 8 & $56.06$ & $32.84$ & $37.61$ & $38.99$ & $34.65$ & $42.58$ & $41.21$ & $\mathbf{47.90}$ \\

\bottomrule
\end{tabular}
}%
\caption{{Performance comparison across aggregation strategies with GLM-4.7-Flash} with varying $K \in \{2,4,8\}$. The same model is used for both rollout and aggregation. \textbf{Bold} indicates the best-performing method per row (excluding Pass). Subscripts report standard deviation across runs. `-' denotes not applicable.}
\label{tab:main_glm}
\end{table}

\begin{table}[t]
\centering
\footnotesize
\resizebox{\linewidth}{!}{%
\setlength{\tabcolsep}{3.8pt}
\begin{tabular}{lrcccccccc}
\toprule
\multirow{2}{*}{\textbf{Task}} & \multirow{2}{*}{$\mathbf K$} & \multirow{2}{*}{\textbf{Pass}} & \multicolumn{4}{c}{\textbf{Heuristic Aggregation}} & \multicolumn{3}{c}{\textbf{LLM-based Aggregation}} \\
\cmidrule(lr){4-7}\cmidrule(lr){8-10}
 & & & \textbf{\MV} & \textbf{\WMV} & \textbf{\BoN} & \textbf{\FewTool} & \textbf{\SolAgg} & \textbf{\SummAgg} & \textbf{\ours} \\
\midrule \midrule

\multirow{4}{*}{BrowseComp}
& 1 & $40.25_{3.26}$ & - & - & - & - & - & - & - \\
& 2 & $50.38_{2.66}$ & $40.74_{2.95}$ & $47.79_{2.82}$ & $48.05_{2.98}$ & $46.21_{2.82}$ & $48.44_{1.00}$ & $\mathbf{49.33}_{2.88}$ & $48.89_{1.13}$ \\
& 4 & $59.53_{1.84}$ & $45.54_{2.13}$ & $54.30_{2.62}$ & $54.83_{2.38}$ & $51.42_{2.22}$ & $55.78_{0.83}$ & $56.00_{0.94}$ & $\mathbf{56.22}_{1.60}$ \\
& 8 & $67.33$ & $47.33$ & $60.00$ & $58.67$ & $56.00$ & $61.33$ & $62.00$ & $\mathbf{62.67}$ \\

\midrule

\multirow{4}{*}{BrowseComp-Plus}
& 1 & $58.67_{2.13}$ & - & - & - & - & - & - & - \\
& 2 & $67.36_{2.07}$ & $58.14_{2.21}$ & $64.93_{2.05}$ & $64.98_{2.06}$ & $64.36_{1.80}$ & $66.00_{1.09}$ & $65.56_{1.13}$ & $\mathbf{66.44}_{1.13}$ \\
& 4 & $73.86_{1.43}$ & $63.44_{1.91}$ & $69.10_{1.37}$ & $69.78_{1.69}$ & $68.54_{1.44}$ & $69.51_{1.13}$ & $\mathbf{71.78}_{0.83}$ & $71.33_{1.89}$ \\
& 8 & $79.33$ & $66.67$ & $69.33$ & $74.00$ & $70.67$ & $73.33$ & $73.33$ & $\mathbf{74.67}$ \\

\midrule

\multirow{4}{*}{HLE}
& 1 & $39.35_{2.39}$ & - & - & - & - & - & - & - \\
& 2 & $50.12_{2.32}$ & $39.19_{2.54}$ & $43.71_{3.21}$ & $43.71_{2.67}$ & $39.26_{2.22}$ & $45.16_{0.53}$ & $45.59_{2.60}$ & $\mathbf{45.81}_{0.53}$ \\
& 4 & $59.87_{1.82}$ & $42.29_{1.96}$ & $45.28_{2.28}$ & $47.43_{2.70}$ & $39.30_{1.79}$ & $49.25_{1.61}$ & $47.96_{0.30}$ & $\mathbf{51.40}_{0.80}$ \\
& 8 & $67.10$ & $43.87$ & $45.16$ & $53.55$ & $41.29$ & $50.97$ & $50.32$ & $\mathbf{54.19}$ \\

\midrule

\multirow{4}{*}{DeepSearchQA}
& 1 & $49.25_{2.27}$ & - & - & - & - & - & - & - \\
& 2 & $60.76_{2.08}$ & - & - & $53.33_{2.23}$ & $52.17_{2.55}$ & $52.67_{2.88}$ & $52.89_{2.06}$ & $\mathbf{55.56}_{1.75}$ \\
& 4 & $70.02_{1.62}$ & - & - & $56.09_{2.21}$ & $53.67_{2.40}$ & $56.00_{1.44}$ & $58.67_{1.09}$ & $\mathbf{59.56}_{1.13}$ \\
& 8 & $76.67$ & - & - & $57.33$ & $54.67$ & $62.67$ & $64.00$ & $\mathbf{66.00}$ \\

\midrule

\multirow{4}{*}{Healthbench-Hard}
& 1 & $12.87_{5.89}$ & - & - & - & - & - & - & - \\
& 2 & $24.52_{3.88}$ & - & - & $13.34_{4.26}$ & $13.62_{3.73}$ & $\mathbf{25.23}_{0.30}$ & $18.84_{1.13}$ & $24.95_{0.39}$ \\
& 4 & $33.45_{2.25}$ & - & - & $14.00_{2.76}$ & $13.70_{2.25}$ & $27.30_{1.91}$ & $22.60_{1.33}$ & $\mathbf{28.01}_{1.06}$ \\
& 8 & $40.74$ & - & - & $13.01$ & $12.83$ & $26.30$ & $23.00$ & $\mathbf{28.06}$ \\

\midrule

\multirow{4}{*}{ResearchRubrics}
& 1 & $40.50_{0.91}$ & - & - & - & - & - & - & - \\
& 2 & $44.77_{0.60}$ & - & - & $40.71_{0.67}$ & $40.02_{0.65}$ & $41.17_{0.42}$ & $34.25_{0.53}$ & $\mathbf{44.49}_{0.52}$ \\
& 4 & $48.11_{0.37}$ & - & - & $40.88_{0.62}$ & $39.77_{0.55}$ & $41.41_{0.59}$ & $36.29_{0.35}$ & $\mathbf{46.61}_{0.42}$ \\
& 8 & $50.97$ & - & - & $42.37$ & $39.58$ & $42.10$ & $37.47$ & $\mathbf{49.36}$ \\

\midrule \midrule

\multirow{4}{*}{\shortstack{\textbf{Average}\\\textbf{(Qwen3.5-122B)}}}
& 1 & $40.15$ & - & - & - & - & - & - & - \\
& 2 & $49.65$ & $40.12$ & $43.18$ & $44.02$ & $42.61$ & $46.45$ & $44.41$ & \textbf{47.69} \\
& 4 & $57.47$ & $42.32$ & $45.22$ & $47.17$ & $44.40$ & $49.88$ & $48.88$ & \textbf{52.19} \\
& 8 & $63.69$ & $43.42$ & $46.19$ & $49.82$ & $45.84$ & $52.78$ & $51.69$ & $\mathbf{55.83}$ \\

\bottomrule
\end{tabular}
}%
\caption{{Performance comparison across aggregation strategies with Qwen3.5-122B}. The table is formatted the same as Table~\ref{tab:main_glm}.}
\label{tab:main_qwen}
\end{table}

\begin{table}[t]
\centering
\footnotesize
\resizebox{\linewidth}{!}{%
\setlength{\tabcolsep}{3.8pt}
\begin{tabular}{lrcccccccc}
\toprule
\multirow{2}{*}{\textbf{Task}} & \multirow{2}{*}{$\mathbf K$} & \multirow{2}{*}{\textbf{Pass}} & \multicolumn{4}{c}{\textbf{Heuristic Aggregation}} & \multicolumn{3}{c}{\textbf{LLM-based Aggregation}} \\
\cmidrule(lr){4-7}\cmidrule(lr){8-10}
 & & & \textbf{\MV} & \textbf{\WMV} & \textbf{\BoN} & \textbf{\FewTool} & \textbf{\SolAgg} & \textbf{\SummAgg} & \textbf{\ours} \\
\midrule \midrule

\multirow{4}{*}{BrowseComp}
& 1 & $50.17_{3.35}$ & - & - & - & - & - & - & -  \\
& 2 & $60.86_{2.32}$ & $50.40_{2.15}$ & $59.40_{2.36}$ & $59.36_{2.34}$ & $57.38_{2.47}$ & $60.00_{1.63}$ & $59.78_{1.13}$ & $\mathbf{60.22}_{1.13}$  \\
& 4 & $68.64_{1.69}$ & $56.58_{2.19}$ & $64.73_{1.88}$ & $66.06_{2.01}$ & $62.63_{1.64}$ & $64.44_{0.63}$ & $66.67_{0.54}$ & $\mathbf{67.33}_{0.0}$  \\
& 8 & $73.33$ & $60.67$ & $66.67$ & $70.67$ & $67.33$ & $70.67$ & $70.00$ & $\mathbf{71.33}$ \\

\midrule

\multirow{4}{*}{BrowseComp-Plus}
& 1 & $64.17_{1.28}$ & - & - & - & - & - & - & - \\
& 2 & $71.12_{1.50}$ & $64.10_{2.02}$ & $69.68_{1.37}$ & $70.02_{1.33}$ & $68.64_{1.69}$ & $70.22_{0.31}$ & $70.44_{0.31}$ & $\mathbf{71.11}_{0.31}$ \\
& 4 & $75.90_{0.94}$ & $67.72_{1.73}$ & $73.57_{1.13}$ & $73.84_{1.11}$ & $71.90_{1.11}$ & $73.56_{0.31}$ & $73.56_{2.45}$ & $\mathbf{74.44}_{0.63}$ \\
& 8 & $78.67$ & $68.67$ & $75.33$ & $76.00$ & $74.67$ & $76.67$ & $76.67$ & $\mathbf{77.33}$ \\

\midrule

\multirow{4}{*}{HLE}
& 1 & $45.73_{1.75}$ & - & - & - & - & - & - & -  \\
& 2 & $56.61_{1.57}$ & $45.14_{1.83}$ & $49.52_{1.94}$ & $49.75_{2.11}$ & $46.77_{2.57}$ & $47.53_{2.38}$ & $49.68_{1.90}$ & $\mathbf{52.90}_{0.91}$  \\
& 4 & $65.28_{1.28}$ & $49.09_{1.70}$ & $52.59_{1.45}$ & $52.76_{1.77}$ & $47.48_{2.26}$ & $50.97_{2.11}$ & {$56.34_{0.80}$} & {$\mathbf{56.99}_{1.22}$}\\
& 8 & $72.26$ & $53.55$ & $57.42$ & $58.06$ & $46.45$ & $54.19$ & $58.06$ & {$\mathbf{60.00}$} \\

\midrule

\multirow{4}{*}{DeepSearchQA}
& 1 & $54.42_{1.66}$ & - & - & - & - & - & - & - \\
& 2 & $65.40_{2.21}$ & - & - & $\mathbf{58.02}_{2.26}$ & $55.90_{1.97}$ & $57.78_{1.57}$ & $57.11_{1.57}$ & {$57.33_{2.18}$} \\
& 4 & $75.04_{1.86}$ & - & - & $61.46_{2.23}$ & $56.50_{1.47}$ & $59.11_{1.91}$ & $61.33_{0.54}$ & $\mathbf{63.78}_{1.13}$ \\
& 8 & $82.67$ & - & - & $64.00$ & $56.00$ & $62.00$ & $61.33$ & $\mathbf{65.33}$ \\

\midrule

\multirow{4}{*}{Healthbench-Hard}
& 1 & $9.67_{3.84}$ & - & - & - & - & - & - & -  \\
& 2 & $21.09_{2.91}$ & - & - & $9.74_{2.38}$ & $9.02_{2.67}$ & $17.30_{1.28}$ & $9.74_{0.91}$ & $\mathbf{19.25}_{1.53}$ \\
& 4 & $30.62_{1.93}$ & - & - & $10.58_{2.06}$ & $8.04_{2.45}$ & $21.25_{1.20}$ & $12.37_{1.32}$ & $\mathbf{24.27}_{0.52}$ \\
& 8 & $39.02$ & - & - & $8.79$ & $5.34$ & $21.84$ & $16.92$ & $\mathbf{24.46}$ \\

\midrule

\multirow{4}{*}{ResearchRubrics}
& 1 & $39.97_{0.71}$ & - & - & - & - & - & - & - \\
& 2 & $43.86_{0.55}$ & - & - & $40.16_{0.55}$ & $39.63_{0.85}$ & $41.88_{0.57}$ & $34.15_{0.55}$ & $\mathbf{43.17}_{0.65}$ \\
& 4 & $47.06_{0.38}$ & - & - & $40.05_{0.58}$ & $39.19_{0.67}$ & $43.26_{0.24}$ & $37.12_{0.42}$ & $\mathbf{44.96}_{0.53}$ \\
& 8 & $49.74$ & - & - & $39.00$ & $38.44$ & $44.00$ & $40.29$ & $\mathbf{45.42}$ \\

\midrule \midrule

\multirow{4}{*}{\shortstack{\textbf{Average}\\\textbf{(MiniMax-M2.5)}}}
& 1 & $44.02$ & - & - & - & - & - & - & - \\
& 2 & $53.16$ & $43.95$ & $47.11$ & $47.84$ & $46.22$ & $49.12$ & $46.82$ & \textbf{50.66} \\
& 4 & $60.42$ & $46.24$ & $49.16$ & $50.79$ & $47.62$ & 52.10 & 51.23 & \textbf{55.30}\\
& 8 & $65.95$ & $47.83$ & $50.58$ & $52.75$ & $48.04$ & $54.90$ & $53.88$ & $\mathbf{57.31}$ \\

\bottomrule
\end{tabular}
}%
\caption{{Performance comparison across aggregation strategies with MiniMax-M2.5}. The table is formatted the same as Table~\ref{tab:main_glm}.}
\label{tab:main_minimax}
\end{table}

\begin{table}[tbp]
\centering
\footnotesize
\resizebox{\linewidth}{!}{%
\setlength{\tabcolsep}{4pt}
\begin{tabular}{lr | *{5}{>{\centering\arraybackslash}p{1.8cm}} | *{4}{>{\centering\arraybackslash}p{1.8cm}}}
\toprule
\multirow{2}{*}{\textbf{Task}} & \multirow{2}{*}{$\mathbf{K}$}
  & \multirow{2}{*}{\parbox{1.8cm}{\centering\textbf{Rollout}\\\textbf{Cost} ($c_r$)}} & \multirow{2}{*}{\parbox{1.8cm}{\centering\textbf{Tool Call}\\\textbf{Cost} ($c_t$)}}
  & \multicolumn{3}{c|}{\textbf{Aggregation Cost ($c_a$)}}
  & \multirow{2}{*}{\parbox{1.8cm}{\centering\textbf{Rollout}\\\textbf{Latency} ($t_r$)}}
  & \multicolumn{3}{c}{\textbf{Aggregation Latency ($t_a$)}} \\
\cmidrule(lr){5-7}\cmidrule(lr){9-11}
 & & & & \textbf{\SolAgg} & \textbf{\SummAgg} & \textbf{\ours}
       & & \textbf{\SolAgg} & \textbf{\SummAgg} & \textbf{\ours} \\
\midrule\midrule

\multirow{4}{*}{BrowseComp}
 & 1 & .0086 & .0390 & - & - & - & 185.10 & - & - & - \\
 & 2 & .0171 & .0780 & .0025 & .0143 & .0044 & 211.82 & 33.97 & 77.72 & 22.67 \\
 & 4 & .0343 & .1560 & .0021 & .0261 & .0048 & 233.74 & 21.38 & 107.27 & 38.51 \\
 & 8 & .0685 & .3119 & .0019 & .0506 & .0047 & 261.41 & 18.69 & 139.42 & 31.65 \\
\midrule
\multirow{4}{*}{BrowseComp-Plus}
 & 1 & .0067 & 0 & - & - & - & 457.47 & - & - & - \\
 & 2 & .0134 & 0 & .0026 & .0142 & .0037 & 536.31 & 42.54 & 84.65 & 25.30 \\
 & 4 & .0268 & 0 & .0021 & .0260 & .0038 & 584.04 & 21.19 & 83.42 & 31.56 \\
 & 8 & .0536 & 0 & .0020 & .0501 & .0038 & 620.15 & 15.31 & 128.79 & 29.88 \\
\midrule
\multirow{4}{*}{HLE}
 & 1 & .0079 & .0190 & - & - & - & 91.77 & - & - & - \\
 & 2 & .0158 & .0379 & .0037 & .0118 & .0046 & 125.04 & 29.41 & 54.56 & 33.55 \\
 & 4 & .0316 & .0758 & .0038 & .0202 & .0053 & 178.32 & 31.07 & 86.67 & 34.19 \\
 & 8 & .0632 & .1517 & .0041 & .0368 & .0055 & 250.64 & 35.24 & 115.87 & 40.32 \\
\midrule
\multirow{4}{*}{DeepSearchQA}
 & 1 & .0068 & .0311 & - & - & - & 105.60 & - & - & - \\
 & 2 & .0135 & .0623 & .0018 & .0113 & .0043 & 159.65 & 14.96 & 55.49 & 30.48 \\
 & 4 & .0271 & .1245 & .0021 & .0210 & .0049 & 225.01 & 18.22 & 69.63 & 33.60 \\
 & 8 & .0542 & .2490 & .0024 & .0401 & .0052 & 261.79 & 21.73 & 105.53 & 36.98 \\
\midrule
\multirow{4}{*}{Healthbench-Hard}
 & 1 & .0037 & .0114 & - & - & - & 41.20 & - & - & - \\
 & 2 & .0075 & .0228 & .0019 & .0081 & .0028 & 52.22 & 22.26 & 46.47 & 23.12 \\
 & 4 & .0150 & .0456 & .0023 & .0146 & .0035 & 67.15 & 22.46 & 58.39 & 25.51 \\
 & 8 & .0299 & .0912 & .0028 & .0276 & .0039 & 88.27 & 26.84 & 81.48 & 27.32 \\
\midrule
\multirow{4}{*}{ResearchRubrics}
 & 1 & .0070 & .0222 & - & - & - & 112.06 & - & - & - \\
 & 2 & .0141 & .0444 & .0028 & .0119 & .0044 & 142.95 & 28.61 & 63.13 & 40.03 \\
 & 4 & .0282 & .0889 & .0034 & .0219 & .0053 & 187.43 & 35.08 & 84.07 & 46.80 \\
 & 8 & .0563 & .1777 & .0045 & .0412 & .0062 & 243.40 & 37.83 & 114.70 & 50.91 \\
\midrule\midrule
\multirow{4}{*}{\shortstack{\textbf{Average}\\\textbf{(GLM-4.7-Flash)}}}
 & 1 & .0068 & .0204 & - & - & - & 165.53 & - & - & - \\
 & 2 & .0136 & .0409 & .0025 & .0119 & .0040 & 204.67 & 28.63 & 63.67 & 29.19 \\
 & 4 & .0272 & .0818 & .0026 & .0216 & .0046 & 245.95 & 24.90 & 81.58 & 35.03 \\
 & 8 & .0543 & .1636 & .0029 & .0411 & .0049 & 287.61 & 25.94 & 114.30 & 36.18 \\

\bottomrule
\end{tabular}
}%
\caption{{Cost and latency breakdown with GLM-4.7-Flash.} All costs are in USD per query, and all latencies are in seconds per query. `-' denotes not applicable at $K\!=\!1$. BrowseComp-Plus incurs no tool cost.}
\label{tab:cost_latency_glm}
\end{table}

\begin{table}[tbp]
\centering
\footnotesize
\resizebox{\linewidth}{!}{%
\setlength{\tabcolsep}{4pt}
\begin{tabular}{lr | *{5}{>{\centering\arraybackslash}p{1.8cm}} | *{4}{>{\centering\arraybackslash}p{1.8cm}}}
\toprule
\multirow{2}{*}{\textbf{Task}} & \multirow{2}{*}{$\mathbf{K}$}
  & \multirow{2}{*}{\parbox{1.8cm}{\centering\textbf{Rollout}\\\textbf{Cost} ($c_r$)}} & \multirow{2}{*}{\parbox{1.8cm}{\centering\textbf{Tool Call}\\\textbf{Cost} ($c_t$)}}
  & \multicolumn{3}{c|}{\textbf{Aggregation Cost ($c_a$)}}
  & \multirow{2}{*}{\parbox{1.8cm}{\centering\textbf{Rollout}\\\textbf{Latency} ($t_r$)}}
  & \multicolumn{3}{c}{\textbf{Aggregation Latency ($t_a$)}} \\
\cmidrule(lr){5-7}\cmidrule(lr){9-11}
 & & & & \textbf{\SolAgg} & \textbf{\SummAgg} & \textbf{\ours}
       & & \textbf{\SolAgg} & \textbf{\SummAgg} & \textbf{\ours} \\
\midrule\midrule

\multirow{4}{*}{BrowseComp}
 & 1 & .0409 & .0315 & - & - & - & 182.79 & - & - & - \\
 & 2 & .0817 & .0629 & .0102 & .0565 & .0213 & 224.26 & 21.96 & 53.60 & 39.07 \\
 & 4 & .1634 & .1258 & .0103 & .1026 & .0309 & 242.48 & 21.31 & 69.14 & 41.04 \\
 & 8 & .3268 & .2517 & .0103 & .1952 & .0209 & 305.38 & 20.93 & 91.82 & 36.25 \\
\midrule
\multirow{4}{*}{BrowseComp-Plus}
 & 1 & .0265 & 0 & - & - & - & 564.68 & - & - & - \\
 & 2 & .0530 & 0 & .0098 & .0544 & .0186 & 690.97 & 25.55 & 58.80 & 36.38 \\
 & 4 & .1060 & 0 & .0099 & .0977 & .0193 & 743.69 & 21.32 & 62.35 & 32.60 \\
 & 8 & .2120 & 0 & .0097 & .1893 & .0182 & 831.50 & 19.99 & 99.29 & 41.32 \\
\midrule
\multirow{4}{*}{HLE}
 & 1 & .0318 & .0149 & - & - & - & 74.53 & - & - & - \\
 & 2 & .0637 & .0298 & .0128 & .0418 & .0173 & 99.80 & 36.98 & 45.68 & 25.04 \\
 & 4 & .1274 & .0597 & .0131 & .0705 & .0197 & 125.14 & 24.30 & 60.16 & 31.51 \\
 & 8 & .2548 & .1194 & .0127 & .1307 & .0203 & 152.97 & 29.85 & 74.72 & 38.28 \\
\midrule
\multirow{4}{*}{DeepSearchQA}
 & 1 & .0277 & .0239 & - & - & - & 89.32 & - & - & - \\
 & 2 & .0553 & .0477 & .0088 & .0437 & .0208 & 110.47 & 16.71 & 48.91 & 39.57 \\
 & 4 & .1106 & .0954 & .0107 & .0776 & .0244 & 129.99 & 22.32 & 54.54 & 38.96 \\
 & 8 & .2212 & .1909 & .0116 & .1494 & .0247 & 173.43 & 20.50 & 65.65 & 40.28 \\
\midrule
\multirow{4}{*}{Healthbench-Hard}
 & 1 & .0080 & .0045 & - & - & - & 22.03 & - & - & - \\
 & 2 & .0161 & .0091 & .0092 & .0215 & .0141 & 26.67 & 24.29 & 32.69 & 26.68 \\
 & 4 & .0321 & .0181 & .0107 & .0376 & .0166 & 29.56 & 26.96 & 40.83 & 28.86 \\
 & 8 & .0642 & .0363 & .0130 & .0680 & .0175 & 35.09 & 30.70 & 53.77 & 32.38 \\
\midrule
\multirow{4}{*}{ResearchRubrics}
 & 1 & .0182 & .0093 & - & - & - & 66.92 & - & - & - \\
 & 2 & .0364 & .0186 & .0183 & .0377 & .0260 & 75.28 & 47.98 & 51.80 & 47.75 \\
 & 4 & .0728 & .0371 & .0213 & .0632 & .0276 & 84.19 & 57.39 & 72.33 & 54.95 \\
 & 8 & .1457 & .0743 & .0261 & .1134 & .0301 & 96.06 & 47.39 & 87.53 & 55.08 \\
\midrule\midrule
\multirow{4}{*}{\shortstack{\textbf{Average}\\\textbf{(Qwen3.5-122B)}}}
 & 1 & .0255 & .0140 & - & - & - & 166.71 & - & - & - \\
 & 2 & .0510 & .0280 & .0115 & .0426 & .0197 & 204.58 & 28.91 & 48.58 & 35.75 \\
 & 4 & .1020 & .0560 & .0127 & .0749 & .0231 & 225.84 & 28.93 & 59.89 & 37.99 \\
 & 8 & .2041 & .1121 & .0139 & .1410 & .0219 & 265.74 & 28.23 & 78.80 & 40.60 \\

\bottomrule
\end{tabular}
}%
\caption{{Cost and latency breakdown with Qwen3.5-122B.} The table is formatted the same as Table~\ref{tab:cost_latency_glm}.}
\label{tab:cost_latency_qwen}
\end{table}

\begin{table}[tbp]
\centering
\footnotesize
\resizebox{\linewidth}{!}{%
\setlength{\tabcolsep}{4pt}
\begin{tabular}{lr | *{5}{>{\centering\arraybackslash}p{1.8cm}} | *{4}{>{\centering\arraybackslash}p{1.8cm}}}
\toprule
\multirow{2}{*}{\textbf{Task}} & \multirow{2}{*}{$\mathbf{K}$}
  & \multirow{2}{*}{\parbox{1.8cm}{\centering\textbf{Rollout}\\\textbf{Cost} ($c_r$)}} & \multirow{2}{*}{\parbox{1.8cm}{\centering\textbf{Tool Call}\\\textbf{Cost} ($c_t$)}}
  & \multicolumn{3}{c|}{\textbf{Aggregation Cost ($c_a$)}}
  & \multirow{2}{*}{\parbox{1.8cm}{\centering\textbf{Rollout}\\\textbf{Latency} ($t_r$)}}
  & \multicolumn{3}{c}{\textbf{Aggregation Latency ($t_a$)}} \\
\cmidrule(lr){5-7}\cmidrule(lr){9-11}
 & & & & \textbf{\SolAgg} & \textbf{\SummAgg} & \textbf{\ours}
       & & \textbf{\SolAgg} & \textbf{\SummAgg} & \textbf{\ours} \\
\midrule\midrule

\multirow{4}{*}{BrowseComp}
 & 1 & .0321 & .0301 & - & - & - & 209.34 & - & - & - \\
 & 2 & .0642 & .0602 & .0055 & .0389 & .0122 & 259.73 & 267.45 & 300.34 & 294.52 \\
 & 4 & .1285 & .1204 & .0049 & .0715 & .0120 & 292.90 & 300.42 & 347.88 & 319.73 \\
 & 8 & .2569 & .2409 & .0049 & .1357 & .0118 & 314.26 & 323.33 & 480.36 & 352.57 \\
\midrule
\multirow{4}{*}{BrowseComp-Plus}
 & 1 & .0298 & 0 & - & - & - & 392.76 & - & - & - \\
 & 2 & .0596 & 0 & .0052 & .0459 & .0114 & 519.48 & 528.28 & 567.93 & 550.86 \\
 & 4 & .1191 & 0 & .0037 & .0876 & .0111 & 632.33 & 639.21 & 708.68 & 663.04 \\
 & 8 & .2382 & 0 & .0025 & .1651 & .0111 & 716.32 & 723.99 & 968.27 & 744.93 \\
\midrule
\multirow{4}{*}{HLE}
 & 1 & .0358 & .0164 & - & - & - & 165.06 & - & - & - \\
 & 2 & .0716 & .0327 & .0088 & .0385 & .0116 & 194.42 & 217.01 & 239.92 & 227.34 \\
 & 4 & .1433 & .0654 & .0087 & .0689 & .0144 & 228.92 & 244.59 & 304.89 & 259.10 \\
 & 8 & .2865 & .1308 & .0082 & .1294 & .0126 & 264.72 & 280.52 & 404.20 & 310.09 \\
\midrule
\multirow{4}{*}{DeepSearchQA}
 & 1 & .0282 & .0268 & - & - & - & 149.91 & - & - & - \\
 & 2 & .0564 & .0536 & .0039 & .0378 & .0141 & 179.42 & 189.97 & 219.72 & 210.96 \\
 & 4 & .1129 & .1072 & .0039 & .0721 & .0147 & 193.99 & 207.52 & 247.52 & 232.83 \\
 & 8 & .2257 & .2144 & .0047 & .1412 & .0148 & 227.26 & 236.98 & 350.82 & 261.86 \\
\midrule
\multirow{4}{*}{Healthbench-Hard}
 & 1 & .0074 & .0052 & - & - & - & 28.86 & - & - & - \\
 & 2 & .0148 & .0105 & .0035 & .0143 & .0057 & 33.13 & 49.61 & 64.43 & 58.19 \\
 & 4 & .0295 & .0209 & .0043 & .0267 & .0070 & 34.08 & 52.90 & 75.92 & 58.94 \\
 & 8 & .0591 & .0419 & .0058 & .0507 & .0084 & 40.55 & 65.48 & 95.72 & 72.02 \\
\midrule
\multirow{4}{*}{ResearchRubrics}
 & 1 & .0151 & .0125 & - & - & - & 93.25 & - & - & - \\
 & 2 & .0302 & .0250 & .0086 & .0257 & .0125 & 109.61 & 159.09 & 161.16 & 166.29 \\
 & 4 & .0603 & .0499 & .0109 & .0480 & .0149 & 122.48 & 177.29 & 192.37 & 177.12 \\
 & 8 & .1207 & .0998 & .0151 & .0919 & .0183 & 146.17 & 203.61 & 270.58 & 213.30 \\
\midrule\midrule
\multirow{4}{*}{\shortstack{\textbf{Average}\\\textbf{(MiniMax-M2.5)}}}
 & 1 & .0237 & .0152 & - & - & - & 173.20 & - & - & - \\
 & 2 & .0474 & .0303 & .0059 & .0335 & .0113 & 215.97 & 235.24 & 258.92 & 251.36 \\
 & 4 & .0949 & .0606 & .0061 & .0625 & .0123 & 250.78 & 270.32 & 312.88 & 285.13 \\
 & 8 & .1898 & .1213 & .0069 & .1190 & .0128 & 284.88 & 305.65 & 428.33 & 325.80 \\

\bottomrule
\end{tabular}
}%
\caption{{Cost and latency breakdown with MiniMax-M2.5.} The table is formatted the same as Table~\ref{tab:cost_latency_glm}.}
\label{tab:cost_latency_minimax}
\end{table}

\begin{table}[t]
\centering
\footnotesize
\setlength{\tabcolsep}{3.8pt}
\resizebox{\linewidth}{!}{%
\begin{tabular}{lr*{7}{w{c}{4.5em}}}
\toprule
\multirow{2}{*}{\textbf{Task}} & \multirow{2}{*}{$\mathbf K$} & \multirow{2}{*}{\textbf{Pass}} & \multicolumn{3}{c}{\textbf{GLM-4.7-Flash}} & \multicolumn{3}{c}{\textbf{MiniMax-M2.5}} \\
\cmidrule(lr){4-6} \cmidrule(lr){7-9}
 & & & \textbf{\SolAgg} & \textbf{\SummAgg} & \textbf{\ours} & \textbf{\SolAgg} & \textbf{\SummAgg} & \textbf{\ours} \\
\midrule \midrule
\multirow{4}{*}{{BrowseComp}}
& 1 & $27.42$ & - & - & - & - & - & - \\
& 2 & $37.86$ & $35.78$ & $34.89$ & $37.78$ & $36.22$ & $37.11$ & \textbf{39.11} \\
& 4 & $49.21$ & $45.33$ & $46.22$ & $45.78$ & $46.22$ & $44.44$ & \textbf{46.44} \\
& 8 & $58.67$ & $53.33$ & $55.33$ & \textbf{56.00} & $51.33$ & $55.33$ & $55.33$ \\

\midrule

\multirow{4}{*}{{BrowseComp-Plus}}
& 1 & $49.08$ & - & - & - & - & - & - \\
& 2 & $59.64$ & $54.67$ & $58.00$ & $58.44$ & $58.22$ & $58.67$ & \textbf{59.33} \\
& 4 & $67.36$ & $63.33$ & $63.33$ & $66.22$ & $64.89$ & $65.33$ & \textbf{68.44} \\
& 8 & $72.00$ & $70.67$ & $70.67$ & $71.33$ & $70.00$ & $71.33$ & \textbf{72.67} \\

\midrule

\multirow{4}{*}{{HLE}}
& 1 & $25.00$ & - & - & - & - & - & - \\
& 2 & $34.01$ & $29.68$ & $30.75$ & \textbf{31.40} & $27.74$ & $30.54$ & $29.46$ \\
& 4 & $42.88$ & $30.97$ & $31.83$ & $32.47$ & $32.26$ & $30.97$ & \textbf{33.76} \\
& 8 & $50.97$ & $32.90$ & $34.84$ & $37.42$ & $35.48$ & $35.48$ & \textbf{38.06} \\

\midrule

\multirow{4}{*}{{DeepSearchQA}}
& 1 & $32.42$ & - & - & - & - & - & - \\
& 2 & $43.55$ & $33.78$ & $38.44$ & $40.44$ & $36.22$ & $35.56$ & \textbf{42.89} \\
& 4 & $54.26$ & $42.44$ & $43.56$ & $42.00$ & $41.11$ & $41.78$ & \textbf{43.78} \\
& 8 & $64.00$ & $46.00$ & $47.33$ & $49.33$ & $48.00$ & $50.00$ & \textbf{52.67} \\

\midrule

\multirow{4}{*}{{Healthbench-Hard}}
& 1 & $8.67$ & - & - & - & - & - & - \\
& 2 & $21.34$ & $10.50$ & $-1.21$ & \textbf{24.45} & $18.06$ & $10.25$ & $22.81$ \\
& 4 & $31.96$ & $13.70$ & $4.76$ & \textbf{25.36} & $22.53$ & $16.85$ & $24.22$ \\
& 8 & $40.91$ & $15.72$ & $7.35$ & \textbf{27.99} & $22.18$ & $18.90$ & $26.74$ \\

\midrule

\multirow{4}{*}{{ResearchRubrics}}
& 1 & $37.47$ & - & - & - & - & - & - \\
& 2 & $42.71$ & $33.84$ & $27.69$ & $41.32$ & $40.43$ & $33.90$ & \textbf{43.42} \\
& 4 & $46.57$ & $37.15$ & $30.83$ & $43.70$ & $43.02$ & \textbf{44.76} & $44.69$ \\
& 8 & $49.79$ & $36.84$ & $31.72$ & $45.31$ & $43.14$ & $40.56$ & \textbf{46.07} \\

\midrule \midrule

\multirow{4}{*}{\textbf{Average}}
& 1 & $30.01$ & - & - & - & - & - & - \\
& 2 & $39.85$ & $33.04$ & $31.43$ & $38.97$ & $36.15$ & $34.34$ & \textbf{39.50} \\
& 4 & $48.71$ & $38.82$ & $36.76$ & $42.59$ & $41.67$ & $40.69$ & \textbf{43.56} \\
& 8 & $56.06$ & $42.58$ & $41.21$ & $47.90$ & $45.02$ & $45.27$ & \textbf{48.59} \\

\bottomrule
\end{tabular}%
}
\caption{Performance of LLM-based aggregation strategies on GLM-4.7-Flash as the base rollout model, comparing GLM (same model) and MiniMax (stronger model) as the LLM aggregator, across varying $K$. \textbf{Bold} indicates the best-performing method per row (excluding Pass). Results show that using a stronger aggregator leads to better performance. `-' denotes not applicable at $K\!=\!1$.}
\label{tab:minimax}
\end{table}

\clearpage
\begin{tcolorbox}[colback=LightOrange,
colframe=orange!90,title=Rollout agent system prompt for agentic search tasks,breakable,listing only,fontupper=\footnotesize]
You are a deep research assistant. Your core function is to conduct thorough, multi-source investigations into any topic. You must handle both broad, open-domain inquiries and queries within specialized academic fields. For every request, synthesize information from credible, diverse sources to deliver a comprehensive, accurate, and objective response. After you have gathered sufficient information, provide the definitive response and complete the task.
\end{tcolorbox}
\captionof{figure}{{Rollout agent system prompt for agentic search tasks, from~\citet{team2025tongyi}.}} 
\label{fig:rollout_qa}
\vspace{15pt}

\begin{tcolorbox}[colback=LightOrange,
colframe=orange!90,title=Additional rollout agent system prompt for deep research tasks,breakable,listing only,fontupper=\footnotesize]
For the given question, please write a comprehensive, evidence-backed answer to scientific questions. You should ground every nontrivial claim in retrieved snippets. Cite using <cite url="...">...</cite> drawn only from returned snippets. Please prefer authoritative sources (peer-reviewed papers, reputable benchmarks/docs) and prioritize recent work for fast-moving areas. You should acknowledge uncertainty and conflicts; if evidence is thin or sources disagree, state it and explain what additional evidence would resolve it. It's important to structure with clear markdown headers and a coherent flow. In each section, write 2-5 sentence paragraphs with clear topic sentences and transitions; use lists sparingly only when they improve clarity. Ideally, you should synthesize rather than enumerate content: it's helpful to group findings across papers, explain relationships, and build a coherent narrative that answers the question, supported by citations. Most importantly, DO NOT invent snippets or citations and never fabricate content. At the end of your response, on a new line, include your confidence score in the following format:
Confidence: \{\{your confidence score between 0\% and 100\% for your answer\}\}
\end{tcolorbox}
\vspace{-5pt}
\captionof{figure}{{Additional rollout agent system prompt for deep research tasks, from~\citet{shao2025dr}.}} 
\label{fig:rollout_dr}
\vspace{15pt}

\begin{tcolorbox}[colback=LightOrange,
colframe=orange!90,title=Rollout agent user message for agentic search tasks,breakable,listing only,fontupper=\footnotesize]
Your response should be in the following format:

Explanation: \{\{your explanation for your final answer\}\}

Exact Answer: \{\{your succinct, final answer\}\}

Confidence: \{\{your confidence score between 0\% and 100\% for your answer\}\}
\end{tcolorbox}
\vspace{-5pt}
\captionof{figure}{{Rollout agent user message for agentic search tasks, from~\citet{wei2025browsecomp}.}} 
\label{fig:rollout_user}
\vspace{15pt}

\begin{tcolorbox}[colback=LightOrange,
colframe=orange!90,title=Additional instruction for evaluating ResearchRubrics,breakable,listing only,fontupper=\footnotesize]
\#\# Handling Negative-Weight Criteria

Some criteria describe **undesirable behaviors** and carry a **negative weight**. The Weight field in the rubric indicates this. For these criteria:

- **Satisfied (Score: 1.0)**: The undesirable behavior IS present in the document. This will reduce the overall score.

- **Not Satisfied (Score: 0.0)**: The undesirable behavior is ABSENT from the document. This is the desired outcome and does not penalize the score.

Your verdict must reflect whether the described behavior is literally present or absent in the document — not whether the document is generally good or bad.
\end{tcolorbox}
\vspace{-5pt}
\captionof{figure}{{Additional instruction for evaluating ResearchRubrics.}} 
\label{fig:eval_rr}

\begin{tcolorbox}[colback=LightOrange,
colframe=orange!90,title=\ours system prompt for agentic search tasks,breakable,listing only,fontupper=\footnotesize]
You are an aggregation agent. You are provided with a task and a set of candidate trajectories from independent agents that attempted to solve it. Your goal is to synthesize the most accurate, complete solution by drawing on the best reasoning and evidence across trajectories.

You do NOT have access to the ground truth solution.

---

**RESPONSIBILITIES**

1. Evaluate tool results and reasoning quality across all candidate trajectories.

2. Identify the most reliable final solution based on verifiable tool observations, logical consistency, and correct tool application.

3. If no single trajectory is fully reliable, synthesize a corrected solution using only verified components from across trajectories.

4. Deliver your synthesized solution in the required format and provide justification.

---

**REQUIRED PROCEDURE**

You must follow these steps before calling `finish'.

1. **Survey the landscape** — Read the TRAJECTORY METADATA in the user message. Identify which trajectories are worth inspecting based on step counts and patterns.

2. **Retrieve full solutions** — Call `get\_solution' (no arguments) to get the final content from every trajectory's last step, or pass a trajectory\_id to retrieve one specific trajectory.

3. **Verify with tool observations** — Do not rely solely on final solutions or a trajectory's own reasoning. For key claims or divergences, go back and inspect what the tools actually returned:
   - Use **search\_trajectory**(trajectory\_id, query) to locate steps where a specific term or claim appears. Use role=`tool' to restrict to actual tool responses when verifying whether a fact was directly observed — this avoids misleading matches on agent reasoning.
   - Use **get\_segment**(trajectory\_id, start\_step, end\_step) to read a contiguous range of steps (max 5). After finding a relevant step via search, read it in full along with surrounding steps to see the raw tool output and surrounding context.

4. **Cross-check** — Confirm: (a) tool observations in the log match what the agent claims, (b) reasoning is not circular, (c) arithmetic and logic are correct.

---

**OPERATIONAL GUIDELINES**

- **Tool results are ground truth; agent reasoning is not.** Within each trajectory, what a tool *returned* is an objective observation. What the agent *concluded* from it is an interpretation that may be wrong. When in conflict, trust the tool output over the agent's written reasoning about it.

- **Count evidence, not trajectories.** A single trajectory with a clear, unambiguous tool observation supporting answer X is stronger evidence than many trajectories that *reasoned* their way to Y without grounding in tool outputs. Majority agreement alone is not sufficient — check what the tools actually showed.

- **Identify Divergence:** Focus on steps where agents disagree. Determine which agent's *observation from the environment* was correct, not which agent sounded more confident.

- **Evidence Grounding:** Ensure tool observations directly support conclusions. If the log shows an error or empty result, the agent cannot validly claim success from that step.

- **Quality over Confidence:** Prefer trajectories with validated, step-by-step reasoning over those that only state a confident conclusion.

---

**COMMON PITFALLS**

- **Hallucinated Observations:** The agent claims a tool returned X, but the log shows Y (or nothing).

- **Silent Failures:** The agent receives an error but continues as if it succeeded.

- **Circular Logic:** The agent assumes the answer before deriving it from data.

- **Arithmetic/Logical Errors:** The data is correct, but the calculation or inference is flawed.

- **Majority Bias:** Do not treat numerical agreement among trajectories as strong evidence. Many trajectories reaching the same conclusion via similar reasoning is weaker than one trajectory with a concrete, verifiable tool result.

---

**SOLUTION FORMAT (finish tool)**

The 'solution' argument must be a single string with exactly two XML sections: <explanation>...</explanation><answer>...</answer>.

- **CORRECT:** Self-contained. A reader who never saw trajectories understands how the answer was derived. No mentions of "trajectory 1", ``get\_solution", or "agent".
  Example:
  <explanation>We need the 2020 population. The census table shows state X with 1.2M and state Y with 0.8M; the question asks for the sum. 1.2 + 0.8 = 2.0 million.</explanation><answer>2.0 million</answer>

- **WRONG:** "Trajectory 2 had the right answer so I chose it." or "According to get\_solution, the answer is 42." Do NOT reference trajectory IDs or tools in the solution.

---

**TERMINATION**

Call `finish' only after verifying key reasoning against actual tool outputs. Do not finish after only reading metadata or only get\_solution; verify at least one critical claim with get\_segment or search\_trajectory when trajectories disagree.
\end{tcolorbox}
\captionof{figure}{{\ours system prompt for agentic search tasks.}} 
\label{fig:aggagent_prompt_qa}
\vspace{20pt}

\begin{tcolorbox}[colback=LightOrange,
colframe=orange!90,title=\ours system prompt for deep research tasks,breakable,listing only,fontupper=\footnotesize]
You are an aggregation agent. You are provided with a task and a set of candidate trajectories from independent agents that attempted to solve it. Your goal is to synthesize the most accurate, complete solution by drawing on the best reasoning and evidence across trajectories.

You do NOT have access to the ground truth.

---

**RESPONSIBILITIES**

1. Evaluate tool results and reasoning quality across all candidate trajectories.

2. Identify the most reliable final solution based on verifiable tool observations, logical consistency, and correct tool application.

3. If no single trajectory is fully reliable, synthesize a corrected solution using only verified components from across trajectories.

4. Deliver your synthesized solution in the required format and provide justification.

---

**REQUIRED PROCEDURE**

You must follow these steps before calling 'finish'.

1. **Survey the landscape** — Read the TRAJECTORY METADATA in the user message. Identify which trajectories are worth inspecting based on step counts and patterns.

2. **Retrieve full solutions** — Call 'get\_solution' (no arguments) to get the final content from every trajectory's last step, or pass a trajectory\_id to retrieve one specific trajectory.

3. **Verify with tool observations** — Do not rely solely on final solutions or a trajectory's own reasoning. For key claims or divergences, go back and inspect what the tools actually returned:
   - Use **search\_trajectory**(trajectory\_id, query) to locate steps where a specific term or claim appears. Use role='tool' to restrict to tool responses when verifying whether a fact was directly observed.
   - Use **get\_segment**(trajectory\_id, start\_step, end\_step) to read a contiguous range of steps (max 5). After finding a relevant step via search, read it in full along with surrounding steps to see the raw tool output and surrounding context.

4. **Cross-check** — Confirm: (a) tool observations in the log match what the agent claims, (b) reasoning is not circular, (c) arithmetic and logic are correct.

5. **Synthesize** — Write a unified response that:
   - Covers every important aspect addressed by any candidate
   - Takes the highest-quality treatment of each aspect (not just the most common)
   - Resolves contradictions by preferring more specific, better-supported, or more precise content
   - Reads as a single coherent response, not a patchwork

---

**QUALITY CRITERIA**

- **Completeness:** The synthesized response must be at least as comprehensive as the best individual candidate, and more comprehensive where candidates complement each other.

- **Accuracy:** Prefer specific, precise content over vague generalizations. When candidates conflict, do not average — choose the more defensible position.

- **Coherence:** The final response must flow naturally. Integrate content rather than concatenating sections.

- **Self-contained:** Do not mention trajectories, agents, candidates, or aggregation anywhere in the response.

- **Citations:** Ground every nontrivial claim in retrieved snippets. Cite using <cite url="...">...</cite> drawn only from returned snippets; never fabricate URLs or content.

---

**COMMON PITFALLS**

- **Cherry-picking the best-sounding candidate:** Length or fluency is not quality. A shorter candidate may cover a critical aspect better.

- **Ignoring minority candidates:** A single candidate covering an important aspect well outweighs many candidates that omit it.

- **Concatenation instead of synthesis:** Stitching sections together without integrating them produces an incoherent response. Rewrite to unify.

- **Contradiction averaging:** If candidates disagree, do not hedge — reason about which is more accurate and commit to it.

- **Omitting details:** If a candidate covers a subtopic with more depth, preserve that depth in the synthesis.

---

**TERMINATION**

Call 'finish' with 'solution\_report' (your complete synthesized response) and 'reason' (a concise account of how you combined the candidates and resolved any conflicts) after you have read and compared all candidates.
\end{tcolorbox}
\vspace{-5pt}
\captionof{figure}{{\ours system prompt for deep research tasks.}} 
\label{fig:aggagent_prompt_dr}
\vspace{15pt}

\begin{tcblisting}{
  colback=LightOrange,
  colframe=orange!90,
  title=\ours tool descriptions,
  breakable,
  listing only,
  listing options={
    language={},
    basicstyle=\ttfamily\footnotesize,
    breaklines=true
  }
}
{
  "type": "function",
  "function": {
    "name": "get_solution",
    "description": "Retrieves the final content from trajectories' last step. Returns a list of {trajectory_id, content} entries.",
    "parameters": {
      "type": "object",
      "properties": {
        "trajectory_id": {
          "type": "integer",
          "description": "Trajectory index. Omit to retrieve all trajectories."
        }
      },
      "required": [],
      "additionalProperties": false
    }
  },
  "strict": true
},
{
  "type": "function",
  "function": {
    "name": "search_trajectory",
    "description": "Searches for keywords or phrases within a single trajectory. Returns top matching steps ranked by relevance score.",
    "parameters": {
      "type": "object",
      "properties": {
        "trajectory_id": {"type": "integer", "description": "Trajectory index to search within."},
        "query": {"type": "string", "description": "Search term or phrase."},
        "role": {"type": "string", "enum": ["tool", "assistant"], "description": "Optional. Filter to 'tool' steps (actual environment observations) or 'assistant' steps only. Omit to search all steps."},
        "k": {"type": "integer", "description": "Max matches to return (default 5, max 10).", "default": 5}
      },
      "required": ["trajectory_id", "query"],
      "additionalProperties": false
    }
  },
  "strict": true
},
{
  "type": "function",
  "function": {
    "name": "get_segment",
    "description": "Reads the full content of a contiguous range of steps from a trajectory (max 5). Use after search_trajectory to inspect a step in full with its surrounding context.",
    "parameters": {
      "type": "object",
      "properties": {
        "trajectory_id": {"type": "integer", "description": "Trajectory index."},
        "start_step": {"type": "integer", "description": "Start step (inclusive)."},
        "end_step": {"type": "integer", "description": "End step (inclusive); end_step - start_step <= 4."}
      },
      "required": ["trajectory_id", "start_step", "end_step"],
      "additionalProperties": false
    }
  },
  "strict": true
},
{
  "type": "function",
  "function": {
    "name": "finish",
    "description": "Submits the final synthesized solution.", 
    "parameters": {
      "type": "object",
      "properties": {
        "solution": {"type": "string", "description": "A comprehensive, standalone solution as a single string with two XML sections: <explanation>detailed reasoning leading to the answer</explanation><answer>the exact answer</answer>. The explanation must be self-contained and make sense without any reference to trajectories or aggregation."},
        "reason": {"type": "string", "description": "Meta-reasoning explaining your decision: how you evaluated the trajectories, what evidence you relied on, and how you resolved any conflicts or inconsistencies."}
      },
    "required": ["solution", "reason"]
    }
  },
  "strict": true
}
\end{tcblisting}
\captionof{figure}{{\ours tool descriptions}.}
\label{fig:aggagent_tools}
\vspace{20pt}

\begin{tcblisting}{
  colback=LightOrange,
  colframe=orange!90,
  title=Variants of \ours finish tool,
  breakable,
  listing only,
  listing options={
    language={},
    basicstyle=\ttfamily\footnotesize,
    breaklines=true
  }
}
Deep Research Tasks:
"solution": {"type": "string", "description": "The synthesized long-form response. Write it as a complete, standalone report -- do not reference trajectories, agents, or aggregation. Cite using <cite url=\"...\">...</cite> if necessary."}

Qwen3.5-122B-A10B on Agentic Search Tasks:
"solution": {"type": "string", "description": "A comprehensive, standalone solution as a single string in the following format:\nExplanation: {detailed reasoning leading to the answer}\nExact Answer: {the exact answer}\nThe explanation must be self-contained and make sense without any reference to trajectories or aggregation."}
\end{tcblisting}
\captionof{figure}{{Variants of \ours finish tool}.}
\label{fig:aggagent_tools_ab}

\end{document}